\documentclass[12pt]{elsarticle}  

\usepackage{graphicx} 
\usepackage{epsfig} 
\usepackage{amsmath} 
\usepackage{amssymb}  
\usepackage{mathtools}
\usepackage{xcolor}
\usepackage[numbers]{natbib}
\usepackage[ruled,vlined]{algorithm2e}
\usepackage{url}
\usepackage{afterpage}
\usepackage{subcaption}
\usepackage{tabu}

\newcommand{\figref}[1]{Figure~\ref{#1}}
\newcommand{\secref}[1]{Section~\ref{#1}}
\newcommand{\tabref}[1]{Table~\ref{#1}}
\renewcommand{\eqref}[1]{Equation~\ref{#1}}

\journal{Image and Vision Computing}

\begin{document}

\begin{frontmatter}

\title{3D Visual Perception for Self-Driving Cars using a Multi-Camera System: Calibration, Mapping, Localization, and Obstacle Detection}

\author[ucb]{Christian H\"{a}ne}
\ead{chaene@eecs.berkeley.edu}
\author[dso]{Lionel Heng}
\ead{lionel\_heng@dso.org.sg}
\author[nus]{Gim Hee Lee}
\ead{gimhee.lee@nus.edu.sg}
\author[graz]{Friedrich Fraundorfer}
\ead{fraundorfer@icg.tugraz.at}
\author[asl]{Paul Furgale}
\ead{paul.furgale@mavt.ethz.ch}
\author[ethz]{Torsten Sattler}
\ead{sattlert@inf.ethz.ch}
\author[ethz,ms]{Marc Pollefeys}
\ead{marc.pollefeys@inf.ethz.ch}

\address[ucb]{Department of Electrical Engineering and Computer Sciences, University of California Berkeley, Berkeley, CA 94720, United States of America}
\address[ethz]{Department of Computer Science, ETH Z\"{u}rich, Universit\"{a}tstrasse 6, 8092 Z\"{u}rich, Switzerland}
\address[dso]{Information Division, DSO National Laboratories, 12 Science Park Drive, Singapore 118225}
\address[nus]{Department of Computer Science, National University of Singapore, 13 Computing Drive, Singapore 117417}
\address[graz]{Institute for Computer Graphics \& Vision, Graz University of Technology, Inffeldgasse 16, A-8010 Graz, Austria}
\address[asl]{Department of Mechanical and Process Engineering, ETH Z\"{u}rich, Leonhardstrasse 21, 8092 Z\"{u}rich, Switzerland}
\address[ms]{Microsoft, One Microsoft Way, Redmond, WA 98052, United States of America}

\begin{abstract}
Cameras are a crucial exteroceptive sensor for self-driving cars as they are low-cost and small, provide appearance information about the environment, and work in various weather conditions.
They can be used for multiple purposes such as visual navigation and obstacle detection. We can use a surround multi-camera system to cover the full 360-degree field-of-view around the car. In this way, we avoid blind spots which can otherwise lead to accidents. To minimize the number of cameras needed for surround perception, we utilize fisheye cameras. Consequently, standard vision pipelines for 3D mapping, visual localization, obstacle detection, etc. need to be adapted to take full advantage of the availability of multiple cameras rather than treat each camera individually. In addition, processing of fisheye images has to be supported. In this paper, we describe the camera calibration and subsequent processing pipeline for multi-fisheye-camera systems developed as part of the V-Charge project. This project seeks to enable automated valet parking for self-driving cars. Our pipeline is able to precisely calibrate multi-camera systems, build sparse 3D maps for visual navigation, visually localize the car with respect to these maps, generate accurate dense maps, as well as detect obstacles based on real-time depth map extraction.
\end{abstract}


\end{frontmatter}

\section{Introduction}
Fully autonomous cars hold a lot of potential: they promise to make transport safer by reducing the number of accidents caused by inattentive or distracted drivers. They can help to reduce emissions by enabling the sharing of a car between multiple persons. They can also make commutes more comfortable and automate the search for parking spots. 
One fundamental problem that needs to be solved to enable full autonomy is the visual perception problem to provide cars with the ability to sense their surroundings. 
In this paper, we focus on the 3D variant of this problem: estimating the 3D structure of the environment around the car and exploiting it for tasks such as visual localization and obstacle detection.

Cameras are a natural choice as the primary sensor for self-driving cars since lane markings, road signs, traffic lights, and other navigational aids are designed for the human visual system.
At the same time, cameras provide data for a wide range of tasks required by self-driving cars including 3D mapping, visual localization, and 3D obstacle detection while working both in indoor and outdoor environments.   
For full autonomy, it is important that the car is able to perceive objects all around it. This can be achieved by using a multi-camera system that covers the full 360$^\circ$ field-of-view (FOV) around the car.  Cameras with a wide FOV, e.g., fisheye cameras, can be used to minimize the number of required cameras, and thus, the overall cost of the system. Interestingly, research in computer vision has mainly focused on monocular or binocular systems. In contrast, limited research is done for multi-camera systems. Obviously, each camera can be treated individually. However, this ignores the geometric constraints between the cameras and can lead to inconsistencies across cameras.

\textcolor{black}{In this paper, we describe a visual perception pipeline that makes full use of a multi-camera system to obtain precise motion estimates and fully exploits fisheye cameras to cover the 360$^\circ$ around the car with as little as four cameras. More precisely, this paper describes the perception pipeline \cite{hane2014real,Heng2014ICRA,Heng2015JFR,leeCVPR13,IROS13_Lee_Robust,LeeICRA14,IROS13_Lee_Structureless,leeISRR13,leeIJRR15,Lee14CVPR,li13,hane2015obstacle} designed for and used in the V-Charge\footnote{\textcolor{black}{Autonomous Valet Parking and Charging for e-Mobility, \url{http://www.v-charge.eu/}}} project \citep{FurgaleIV2013}. Given the amount of work required to implement such a pipeline, it is clear that the description provided in this paper cannot cover all details. Instead, this paper is intended as an overview over our system that highlights the important design choices we made and the fundamental mathematical concepts used in the approach. As such, we cover the calibration of multi-camera systems \cite{Heng2014ICRA,li13}, including the extrinsic calibration of each camera with respect to the wheel odometry frame of the car, the mathematical models for ego-motion estimation of a multi-camera system \cite{leeCVPR13,Lee14CVPR}, as well as Simultaneous Localization and Mapping (SLAM) \cite{IROS13_Lee_Robust,LeeICRA14,IROS13_Lee_Structureless} and visual localization \cite{leeISRR13,leeIJRR15} for multi-camera systems. In addition, we discuss depth map estimation from fisheye images \cite{hane2014real} and how to obtain dense, accurate 3D models by fusing the depth maps, efficient re-calibration using existing SLAM maps \cite{Heng2015JFR}, and real-time obstacle detection with fisheye cameras \cite{hane2015obstacle}. We provide references to the original publications describing each part in detail.}

\textcolor{black}{To the best of our knowledge, ours is the first purely visual 3D perception pipeline based on a multi-camera system and the first pipeline to fully exploit fisheye cameras with little to no visual overlap. Given the advantages of such a pipeline, and its inherent complexity, we believe that such an overview is fundamentally important and interesting for both academia and industry alike. Besides this overview, which is the main contribution of the paper, we also describe our dense height map fusion approach, which has only been published in an earlier version for indoor environments before.}

In the following sections, we give a brief overview of our pipeline and review existing perception pipelines for self-driving cars. 

\subsection{System Overview}
\label{sec:v_charge}

The 3D visual perception pipeline described in this paper was developed as part of the V-Charge project, funded by the EU's Seventh Framework Programme. The goal of the project was to enable fully autonomous valet parking and charging for electric cars. 
As indoor parking garages were a major target, our 3D perception pipeline does not use any GPS information.

\begin{figure}[t]
  \centering
  \includegraphics[width=0.9\textwidth]{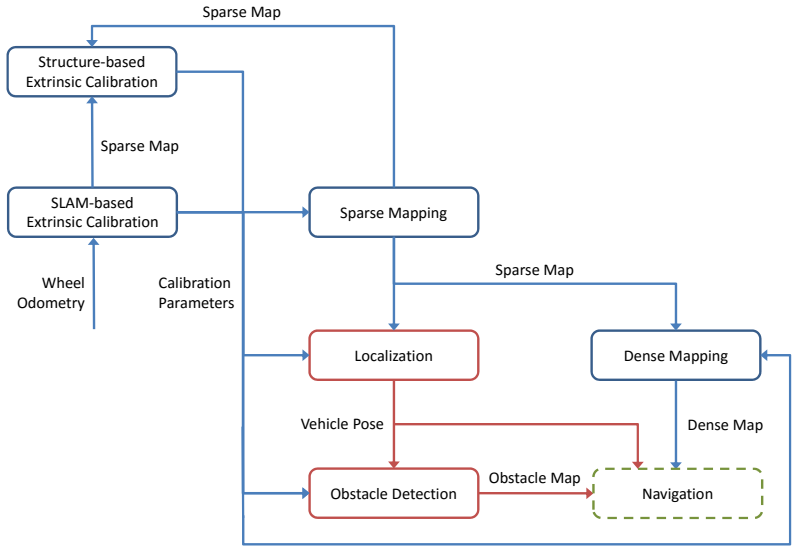}
  \caption{Our 3D visual perception pipeline from calibration to mapping. Each component in the pipeline is marked with a solid outline subscribes to images from the multi-camera system. Components marked with blue outlines run offline while those with red outlines run online. The outputs from our pipeline: the vehicle pose, obstacle map, and dense map, can be used for autonomous navigation.}
  \label{fig:pipeline}
\end{figure}

\figref{fig:pipeline} provides an overview of our pipeline. Given sequences of images recorded for each camera during manual driving and the corresponding wheel odometry poses, our SLAM-based calibration approach, described in \secref{sec:calibration}, computes the extrinsic parameters of the cameras with respect to the vehicle's wheel odometry frame.
The extrinsic parameters are then used by all the other modules in the pipeline. Our sparse mapping module, described in \secref{sec:sparse_mapping}, estimates the ego-motion of the car from 2D-2D feature matches between the camera images. The estimated motion is then used to build a sparse 3D map. 
The sparse 3D map is used by our localization method, described in \secref{sec:Localization}, to estimate the position and orientation of the car with respect to the 3D map from 2D-3D matches between features in the camera images and 3D points in the map. Given the poses estimated by the sparse mapping module, our dense mapping module, described in \secref{sec:denseMapping}, estimates a dense depth map per camera image and fuses them into an accurate 3D model. \secref{sec:applications} describes structure-based calibration and obstacle detection methods, both of which leverage our pipeline. Our structure-based calibration approach uses the sparse 3D map for efficient calibration while our obstacle detection uses camera images and the vehicle pose estimates from the localization to build an obstacle map.

\begin{figure}[t]
  \centering
  \includegraphics[width=0.6\textwidth]{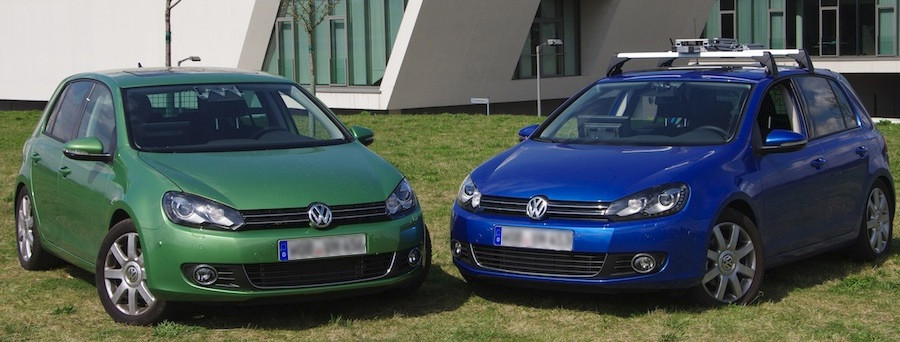}%
  \includegraphics[width=0.39\textwidth]{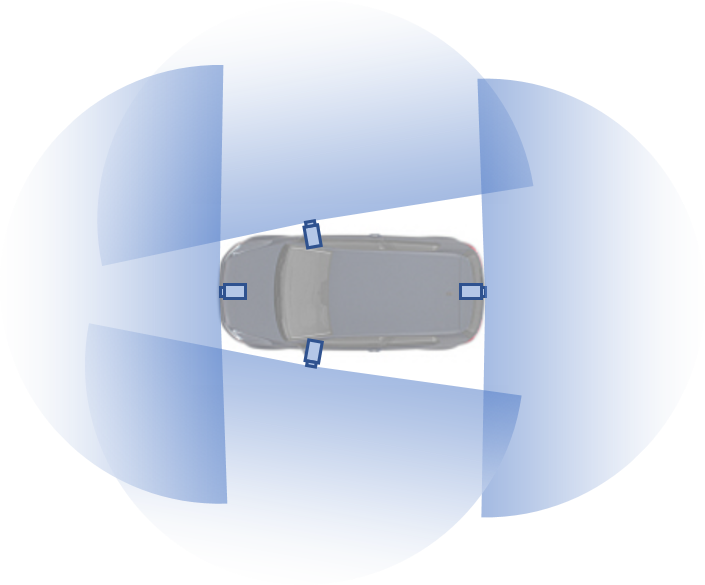}%
  \caption{(left) The two cars from the V-Charge project. 
  (right) The cameras are mounted in the front and back and in the side view mirrors.}
  \label{fig:cars}
\end{figure}

The platform used by the V-Charge project is a VW Golf VI car modified for vision-guided autonomous driving. 
As shown in \figref{fig:cars}, four fisheye cameras are used to build a multi-camera system. 
Each camera has a nominal FOV of 185\ensuremath{^\circ} and outputs 1280 $\times$ 800 images at 12.5 frames per second (fps). The cameras are hardware-synchronized with the wheel odometry of the car. 

\subsection{Related Work}
Many major car manufacturers, technology companies such as Google, and universities are actively working on developing self-driving cars.  Google's self-driving car project \citep{google-selfdriving-car} relies on a combination of lasers, radars, and cameras in the form of a roof-mounted sensor pod to navigate pre-mapped environments. The cars use a custom-built, expensive and omnidirectional 3D laser to build a laser reflectivity map \citep{LevinsonICRA2010} of an environment, localize against that map, and detect obstacles. In contrast, we use a low-cost surround view multi-camera system. 
Such multi-camera systems can be found on mass-market cars from well-known automotive manufacturers,  including BMW, Infiniti, Land Rover, Lexus, Mercedes-Benz, and Nissan. 

The autopilot feature \citep{tesla-autopilot} in cars produced by Tesla Motors enables autonomous driving on highways through a combination of a forward-looking camera and radar. Due to the limited field-of-view, these cars cannot execute autonomous driving in cities which require surround perception. In contrast, our multi-camera system is designed to offer a surround view around the car and with no blind spots.

Similarly, the Bertha Benz self-driving car \citep{ZieglerITSM2014} uses vision and radar albeit with many more sensors to achieve a near-surround-view. The car uses front and rear stereo cameras with a $44^{\circ}$ FOV to build a sparse map for localization. Additional to the stereo cameras the car is equipped with separate rear and front monocular cameras with a $90^{\circ}$ FOV, which are used for localization itself. Notice that this map does not include visual features observed from the sides of the car, and localization may fail if the front and rear cameras are occluded.
In addition, the two cameras used for localization provide two independent pose estimates which are fused in a filter-based framework. In contrast, we use all cameras jointly by treating the camera system as a generalized camera and developing novel algorithms for both, sparse mapping and localization. 

\section{Calibration}
\label{sec:calibration}
Calibration is an essential prerequisite for the use of multi-camera systems in autonomous driving. Projection of a 3D scene point to an image point requires the knowledge of both intrinsic and extrinsic parameters for the multi-camera system in addition to the vehicle pose. In our case, the intrinsic and extrinsic parameters associated with each camera correspond respectively to the parameters of a chosen camera model and that camera's 6-DoF pose with respect to the vehicle's odometry frame.

An imprecise calibration impacts all parts of a visual perception pipeline, leading to inaccurate mapping, localization, and obstacle detection results. 
Furthermore, environmental changes, wear and tear, and vibrations cause calibration parameters to slowly deviate from their true values over time. Multi-camera-based algorithms are sensitive to such calibration errors, and hence, frequent re-calibration is needed. Thus, we developed automatic calibration methods for the car's multi-camera system, which is capable of estimating accurate calibration parameters.

The vast majority of existing work \citep{kumar08,lebraly11,li13} on calibration of multi-camera systems requires a fiducial target, usually a pattern board. The fiducial target contains landmarks with known 3D coordinates which allows target-based calibration methods to compute the inter-camera transformations with metric scale. Our SLAM-based calibration \citep{Heng2015JFR} removes the need for a fiducial target by relying on features naturally occurring in the environment instead. This calibration jointly estimates the inter-camera transformations and builds a sparse map of the environment. Metric scale is inferred from wheel odometry data. Only one other SLAM-based calibration method \citep{carrera11} for multi-camera systems exists but only estimates the inter-camera transformations up to scale.

In this section, we describe the camera model \citep{Mei2007ICRA} that is used throughout this paper, and our SLAM-based extrinsic calibration which does not assume overlapping fields of view for any pair of cameras. The code for both the intrinsic and extrinsic calibration is publicly available
\footnote{\url{https://github.com/hengli/camodocal}}.

\subsection{Camera Model}
\label{sec:cameraModel}
\begin{figure}[t]
  \centering
  \includegraphics[width=0.6\textwidth]{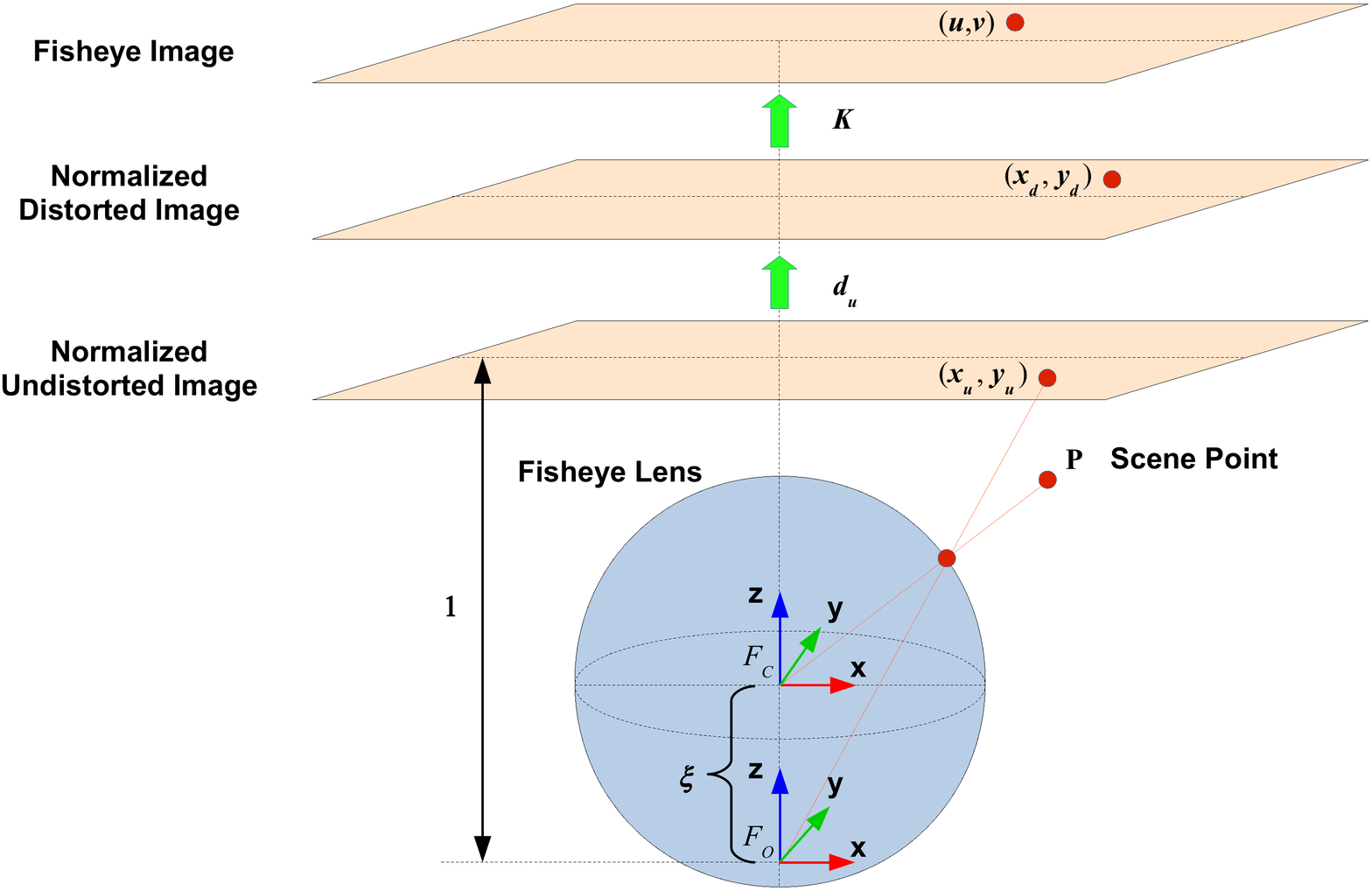}
  \caption{We use the camera model from \citep{Mei2007ICRA}, which comprises both a unified projection model and a radial-tangential distortion model.}
  \label{fig:camera_model}
\end{figure}

\figref{fig:camera_model} illustrates how a scene point is reprojected onto an fisheye image using the camera model from \citep{Mei2007ICRA}, which we explain in the following. For a scene point with coordinates $\mathbf{P}_{F_C} = [X\ Y\ Z]^T$ in the camera's reference frame $F_C$, its projection onto a unit sphere, transformed into another reference frame $F_O$ via a $z$-translation with magnitude $\xi$, is given by
\begin{equation}
  \mathbf{P}_{F_O} = \frac{\mathbf{P}_{F_C}}{\|\mathbf{P}_{F_C}\|} + \begin{bmatrix}0 & 0 & \xi\end{bmatrix}^T \enspace, 
\end{equation}
where $\xi$ denotes the mirror parameter in the unified projection model.

We reproject $\mathbf{P}_{F_O}$ onto the normalized undistorted image plane, and compute the coordinates $(x_u,y_u)$ of the resulting image point as
\begin{equation}
  \begin{bmatrix} x_u & y_u & 1 \end{bmatrix}^T = \frac{\mathbf{P}_{F_O}}{z_{\mathbf{P}_{F_O}}} \enspace, 
\end{equation}
where $z_{\mathbf{P}_{F_O}}$ is the $z$-component of $\mathbf{P}_{F_O}$.

The camera model from \citep{Mei2007ICRA} models the radial and tangential distortions caused by an imperfect lens by computing the coordinates $(x_d,y_d)$ of the normalized distorted image point as
\begin{equation}
  \begin{aligned}
    r &= x_u^2 + y_u^2 \enspace,\\
    \mathbf{d}_u &= \begin{bmatrix} x_u(k_1 r + k_2 r^2) + 2 p_1 x_u y_u + p_2(r + 2 x_u^2) \\ y_u(k_1 r + k_2 r^2) + p_1(r + 2 y_u^2) + 2 p_2 x_u y_u \end{bmatrix} \enspace,\\
    \begin{bmatrix} x_d & y_d \end{bmatrix}^T &= \begin{bmatrix} x_u & y_u \end{bmatrix}^T + \mathbf{d}_u \enspace .
  \end{aligned}
\end{equation}
Here, $[k_1,k_2]$ and $[p_1,p_2]$ are the radial and tangential distortion parameters. 

Finally, the coordinates $(u,v)$ of the image point in the fisheye image are obtained by applying the projection matrix $\mathbf{K}$: 
\begin{equation}
  \begin{aligned}
    \mathbf{K} &= \begin{bmatrix} \gamma_1 & 0 & u_0 \\ 0 & \gamma_2 & v_0 \\ 0 & 0 & 1 \end{bmatrix} \enspace,\\
    \begin{bmatrix} u & v & 1 \end{bmatrix}^T &= \mathbf{K} \begin{bmatrix} x_d & y_d & 1\end{bmatrix}^T \enspace, 
  \end{aligned}
\end{equation}
where $[\gamma_1,\gamma_2]$ and $[u_0,v_0]$ are the focal lengths and principal point coordinates respectively in the unified projection model.

We use an automatic chessboard-based method for the intrinsic camera calibration which is described in \citep{Mei2007ICRA}. We typically obtain an average reprojection error of 
{\raise.17ex\hbox{$\scriptstyle\mathtt{\sim}$}}0.2 pixels for each camera.

\subsection{SLAM-based Extrinsic Calibration}
\label{sec:slam_extrinsic_calibration}
\begin{figure}
  \centering
  \includegraphics[width=0.5\textwidth, angle = -90, trim = 20mm 20mm 30mm 50mm, clip]{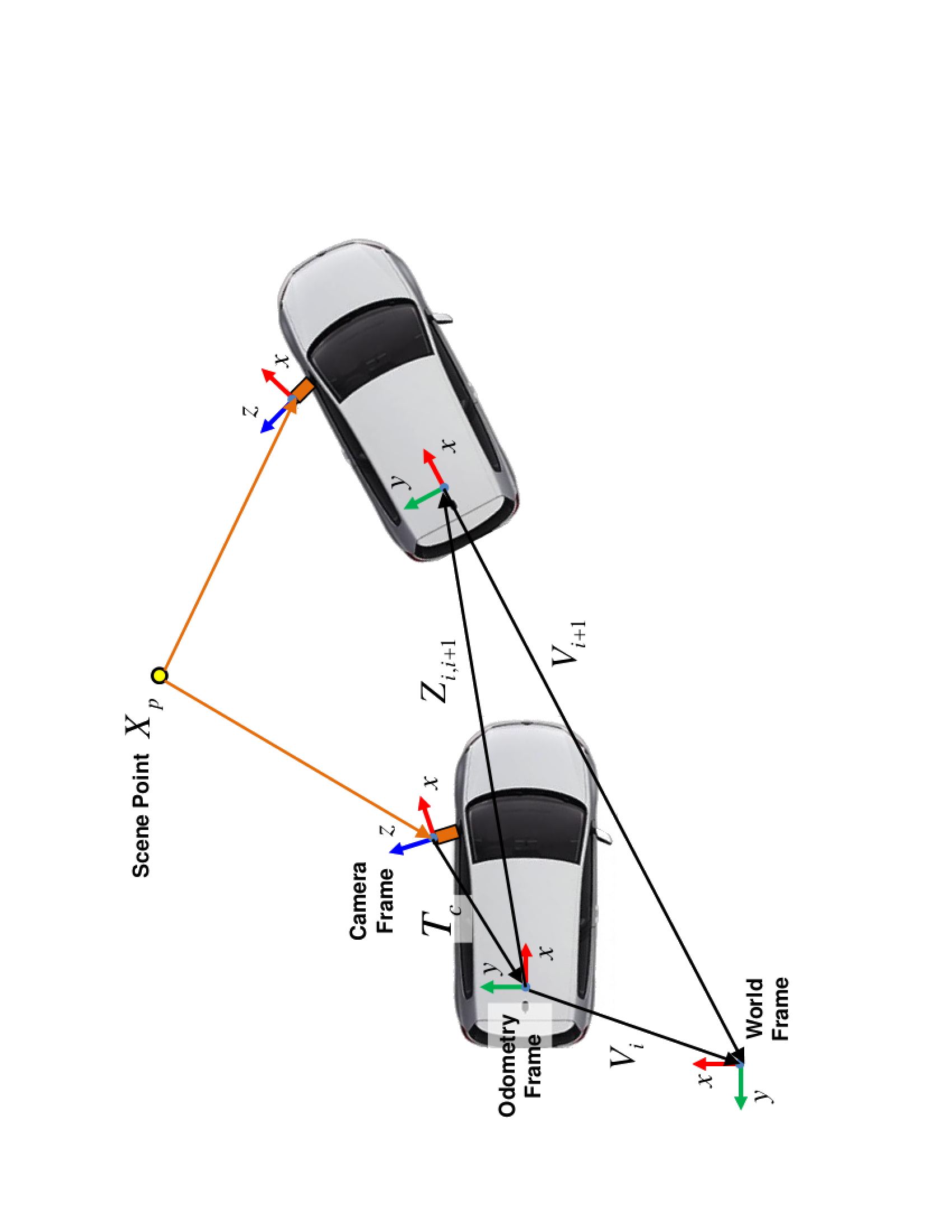}
  \caption{Visualization of the different reference frames and variables described in the NLLS problem which our SLAM-based extrinsic calibration has to solve.}
  \label{fig:frames_slamcalib}
\end{figure}

Given the intrinsic calibration for each camera, our SLAM-based extrinsic calibration method relies on features naturally occurring in the environment and the motion of the car rather than using a fiducial target. We use  OpenCV's SURF \cite{Bay2008} implementation to extract features and their descriptors from images. Essentially, our calibration approach finds a solution to the non-linear least squares (NLLS) problem comprising two sets of inverse-covariance-weighted residuals:
\begin{equation}
 \label{eq:joint_opt}
 \min_{\mathbf{T}_c, \mathbf{V}_i, \mathbf{X}_p} \quad \sum_{c,i,p} \mathbf{\rho} \left( {\Delta \mathbf{z}_{c,i,p}}^T \mathbf{W}_1 \Delta \mathbf{z}_{c,i,p} \right)
 + \sum_{i} {\Delta \mathbf{u}_i}^T \mathbf{W}_2 \Delta \mathbf{u}_i \enspace ,
\end{equation}
where
\begin{equation}
\begin{aligned}
\Delta \mathbf{z}_{c,i,p} &= \boldsymbol\pi(\mathbf{T}_{c}, \mathbf{V}_{i}, \mathbf{X}_p) - \mathbf{p}_{c i p} \enspace, \\
\Delta \mathbf{u}_i &= \mathbf{h}(\mathbf{V}_i, \mathbf{V}_{i+1}) - \mathbf{Z}_{i,i+1} \enspace .
\end{aligned}
\end{equation}
The first and second sets of residuals correspond to the sum of squared image reprojection errors and the sum of squared relative pose errors. $W_1$ and $W_2$ denote the inverse of the measurement covariance corresponding to the first and second sets of residuals, respectively. 
$\boldsymbol\pi$ is a projection function that computes the image coordinates
of the scene point $\mathbf{X}_p$ seen in camera $c$ given the vehicle pose $\mathbf{V}_i$, and the rigid body transformation from the camera
frame to the odometry frame $\mathbf{T}_c$. $\mathbf{p}_{cip}$ is the observed image coordinates
of $\mathbf{X}_p$ seen in camera $c$ with the corresponding vehicle pose $\mathbf{V}_i$.
$\mathbf{\rho}$ is a robust cost function used for minimizing the influence
of outliers. $\mathbf{h}$ is a function that computes the relative pose given two
consecutive vehicle poses $\mathbf{V}_i$ and $\mathbf{V}_{i+1}$,
and $\mathbf{Z}_{i,i+1}$ is the observed relative pose between
$\mathbf{V}_i$ and $\mathbf{V}_{i+1}$ as measured from wheel odometry data. 
We visualize the reference frames and variables in \figref{fig:frames_slamcalib}.

This problem is highly non-convex, and thus, contains many local minimas. Hence, a good initialization is required, which we obtain through a series of steps described below.

\subsubsection{Camera-Odometry Calibration}
The first step obtains an initial estimate of the extrinsic parameters. In this step, we find a least-squares solution to the planar hand-eye calibration problem \citep{guo12} that
relates the relative motions of the camera and the vehicle, and is
not degenerate under planar motion. The relative motions of the camera and the vehicle are estimated via monocular visual odometry and wheel odometry, respectively. The solution yields a 5-DoF transformation between the camera and odometry frames; the height of the camera with respect to the odometry frame is not observable due to planar motion.

\subsubsection{Scene Point Reconstruction}
The second step refines the extrinsic parameters based on a 3D map of the scene: For each camera, we use the initial estimate of the extrinsic parameters and the vehicle poses, which are set to the odometry poses, to triangulate the 2D-2D feature correspondences found from monocular visual odometry. We then jointly refine the extrinsic parameters and scene point coordinates by minimizing the cost function in \eqref{eq:joint_opt} while fixing the vehicle poses.

The estimates of the extrinsic parameters, vehicle poses, and scene point coordinates obtained up to this point  are still inaccurate due to two issues: The wheel odometry data used to initialize the vehicle poses is inaccurate both locally and globally. The 1 cm resolution of the wheel encoders causes the local inaccuracy while the global inaccuracy arises from the pose drift due to the interoceptive nature of the wheel odometry. The second issue is that there are no feature correspondences between different cameras and thus no constraints that relate the extrinsic parameters of the different cameras in our multi-camera system.

We handle the first issue by finding loop closures and performing pose graph optimization to correct the drift inherent in wheel odometry. These loop closures yield inter-camera feature correspondences, but they are not useful priors for estimating the extrinsic parameters as these feature correspondences often correlate different cameras observing the same scene points far apart in time. In order to accurately estimate the extrinsic parameters, we require inter-camera correspondences that correlate to features found in different cameras observing the same scene points close in time. These correspondences provide much stronger constraints.

\subsubsection{Inter-Camera Feature Matching}
In our setup, the cameras are placed to have minimal overlapping fields of view \textcolor{black}{to minimize the number of cameras and thus production costs}. 
For inter-camera feature matching, we maintain a local frame history for each camera and do exhaustive  feature-matching between all frames from each camera and all frames stored in the local frame history corresponding to the other cameras. Obviously, longer local frame histories lead to longer matching times. 

In our implementation, the local frame history for each camera
spans a distance of 3 meters as measured from the vehicle poses. 

\begin{figure}
  \centering
  \includegraphics[width=\linewidth]{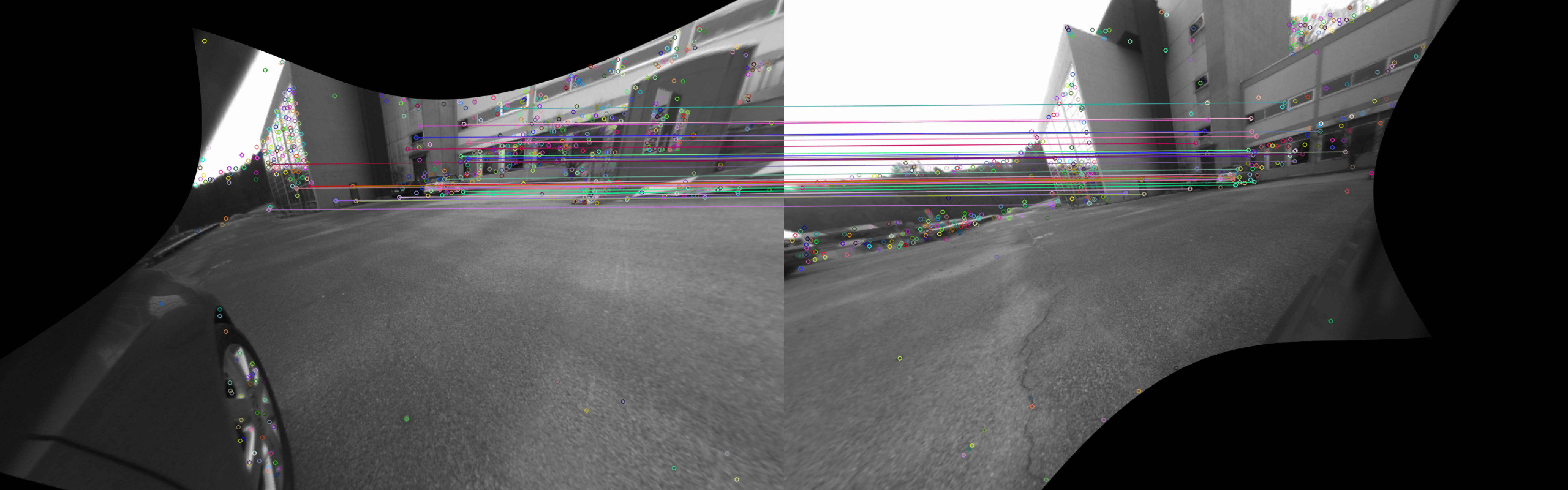}
  \caption{Inlier feature point correspondences between rectified pinhole images
                in the right and front cameras.}
  \label{fig:inlier_matches_rectified}
\end{figure}

With significant viewpoint differences between different cameras at different times, feature matching may yield only few correspondences and even less correct matches. The large distortion effects inherent to our fisheye cameras further aggravate this problem. 
As shown in \figref{fig:inlier_matches_rectified}, we thus rectify these two images on a common image plane before feature extraction to find more inlier matches. 
The plane corresponds to the average rotation between the first camera's pose
and the second camera's pose; the camera poses are computed using the current estimates of the 
extrinsics and the vehicle poses.

\subsubsection{Joint Optimization}
\begin{figure*}
  \centering
  \includegraphics[width=0.49\textwidth]{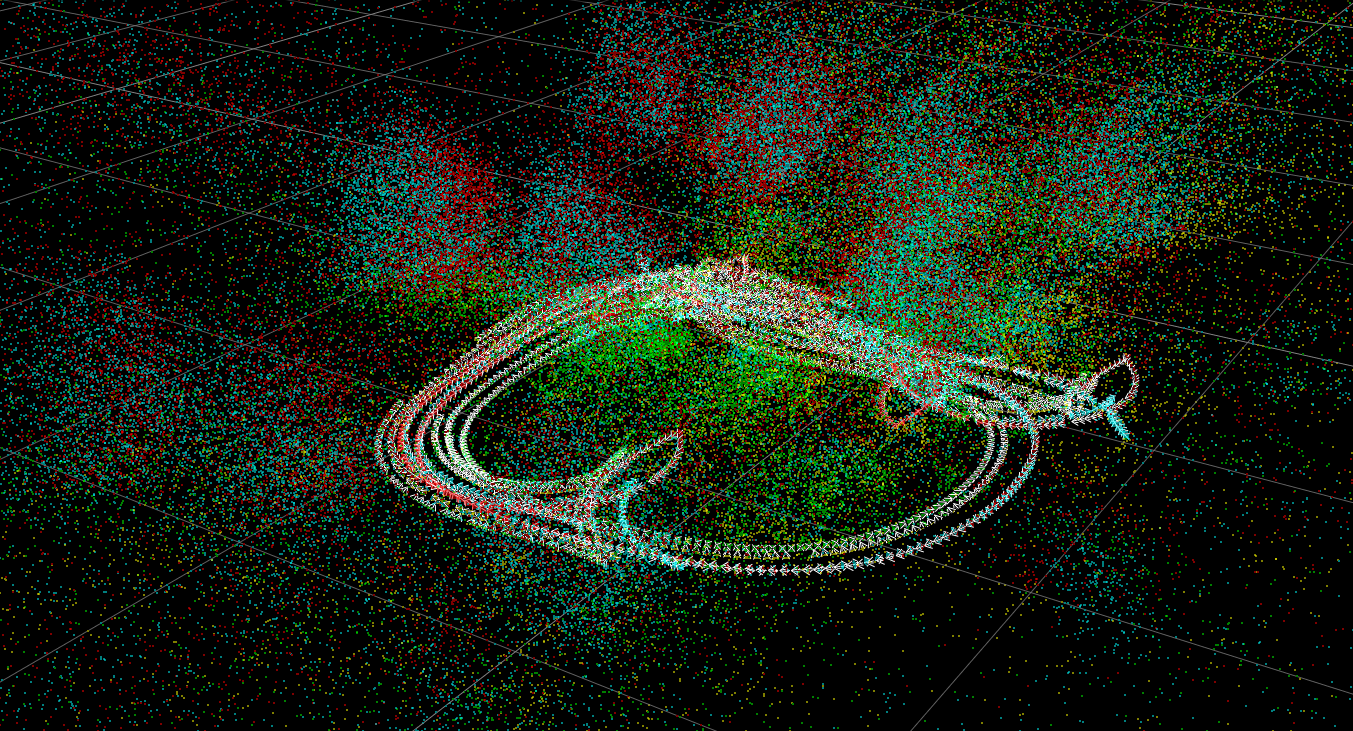}
  \includegraphics[width=0.49\textwidth]{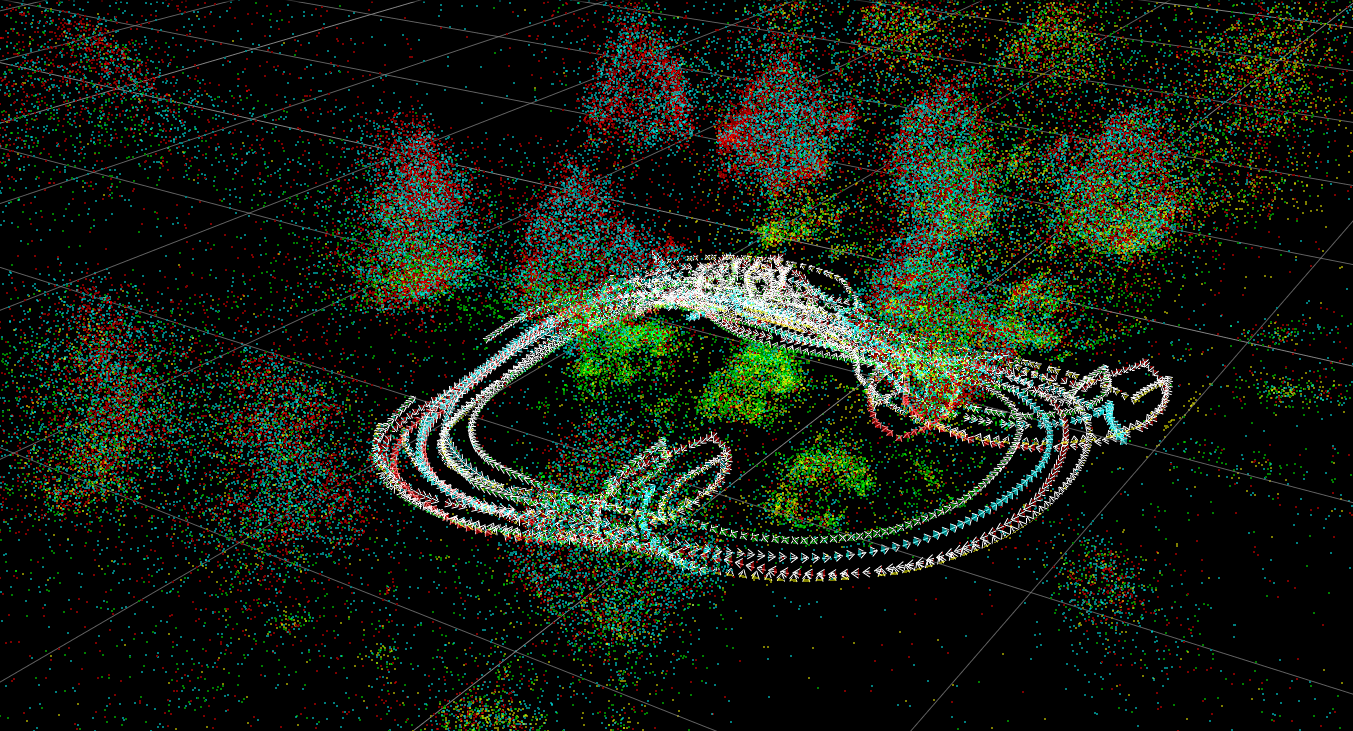}
  \caption{Sparse maps generated with (left) the initial parameters from intrinsic and hand-eye calibration and (right) the parameters obtained after optimization.
           Points seen by a certain camera are assigned an unique color.}
  \label{fig:calibration}
\end{figure*}

Now that we have a good initial estimate of every parameter, we are able to solve \eqref{eq:joint_opt} using a NLLS solver. We obtain an accurate estimate of the extrinsic parameters for the multi-camera system with the help of strong priors provided by inter-camera feature correspondences found in the previous feature-matching step. \figref{fig:calibration} compares two sparse maps;
the map on the left is generated with parameters initially estimated from the steps prior to the joint optimization, while the map on the right is generated
with the optimized parameters from the joint optimization. This figure qualitatively shows that the optimized 3D points, and in turn, the extrinsic parameters are significantly more accurate as the corresponding sparse map is much more well-defined.

\subsubsection{Summary}
More details on the SLAM-based extrinsic calibration and quantitative evaluations of its accuracy can be found in our original publication \citep{Heng2015JFR}. Our SLAM-based extrinsic calibration allows us to calibrate a multi-camera system anywhere, in an unsupervised manner, and without the need for initial guesses of the extrinsic parameters and overlapping fields of view. At the same time, our calibration generates a sparse map of the environment which can be used for visual localization.
\section{Sparse Mapping}
\label{sec:sparse_mapping}

The calibration method described in the last section requires that enough features are found in each camera image. Thus, we usually perform calibration in small, highly textured scenes. In order to be able to also map larger, less textured scenes, we adapted a standard visual SLAM / Structure-from-Motion (SfM) \cite{Hartley2004} framework to process the images collected from our multi-camera system. Like the calibration described above also our sparse mapping system uses SURF \cite{Bay2008} features. Our approach exploits the known camera calibration and is thus more efficient at building sparse maps offline than the SLAM-based approach introduced in the last section. In the following, we describe how to estimate the ego-motion of the camera system (\secref{sec:sparse_mapping:ego_motion}) and how to detect and handle loop-closures (\secref{sec:sparse_mapping:loop_closure}).

\subsection{Ego-Motion Estimation}
\label{sec:sparse_mapping:ego_motion}
The first step toward building a sparse map is to estimate the ego-motions of the multi-camera system based on the images collected during driving. 
A na\"{i}ve 
approach for ego-motion estimation would be to treat each camera individually and compute the relative motion between consecutive frames using classical epipolar geometry \cite{Hartley2004}. However, the resulting motion estimated for each camera is not guaranteed to be consistent with the motions of the other cameras. 
We thus compute the ego-motion jointly for all cameras by modeling our multi-camera system as a generalized camera \cite{Pless03}. 

The main difference between a generalized and a perspective camera is that light rays passing 
through the 2D image features and their respective camera centers do not meet at a single center of projection.
Pless $et.~al.$ \cite{Pless03} suggested that the light rays from a generalized camera can be expressed in a common reference frame with the Pl\"{u}cker line representation. 

\begin{figure}[t]
  \centering
      \includegraphics[width=0.5\columnwidth]{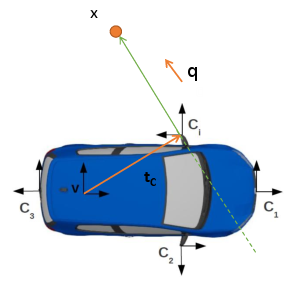}
    \caption{Illustration of a Pl\"{u}cker light ray.}
\label{img:Plucker}
\end{figure}

As illustrated in \figref{img:Plucker}, let $V$ denote the reference frame of the multi-camera system and let $(R_C, t_C)$ be the pose of camera $C$ in $V$. A Pl\"{u}cker line expressed in $V$ 
is given by $l = [q^T~q'^T]^T$, with $q = R_C \hat{x}$ 
and $q' = t_C \times q$. Here, $\hat{x}= \bf{K}^{-1}x$ is the viewing ray corresponding to a 2D point $x$ in the image of camera $C$. 

Given a pair of image point correspondence expressed as Pl\"{u}cker lines $l \leftrightarrow l'$ over two consecutive frames $V$ and $V'$, the Generalized Epipolar Constraint (GEC) \cite{Pless03,Li08} is given 
by 
\begin{equation}
 l'^T \underbrace{\begin{bmatrix} E & R \\ R & 0 \end{bmatrix}}_{E_{GC}}
 l = 0 \enspace ,
\label{eq:gec}
\end{equation}
where $E=\lfloor t \rfloor_{\times} R$ is the conventional Essential matrix \cite{Hartley2004}, and $(R,t)$ is the unknown relative motion between $V'$ and $V$.
$E_{GC}$ is the  
generalized Essential matrix.
 Note that the absolute metric scale for $t$ can be obtained from the GEC since the point correspondences expressed as Pl\"{u}cker lines
contain metric scale information from the extrinsics.

Using \eqref{eq:gec}, $(R,t)$ can be computed from 17 correspondences by solving a linear system 
\cite{Pless03,Li08}. However, this 17-point solver is not practical for real-world applications, where it is necessary to robustly handle wrong correspondences inside a RANSAC loop \cite{Fischler81}. 
In order to find the correct model with a probability of $99\%$, RANSAC estimate and evaluate  
$m=\frac{\ln(1-0.99)}{\ln(1-v^n)}$ pose hypotheses from random samples of size $n$. 
Here, $v$ is the inlier ratio, i.e., the ratio of correct matches among all correspondences. 
For $v=0.5$, the 17-point algorithm ($n=17$) needs $603606$ iterations. 
The non-linear 6-point algorithm by Stew{\'e}nius $et~al.$~\cite{Stewenius05} only requires 292 iterations, but also generates up to 64 solutions that need to be considered in each iteration. 
As a result, both solvers are not applicable in practical scenarios where fast RANSAC run-times are required. 

\begin{figure}[t]
  \centering
      \includegraphics[width=0.5\columnwidth,trim = 35mm 105mm 35mm 35mm, clip]{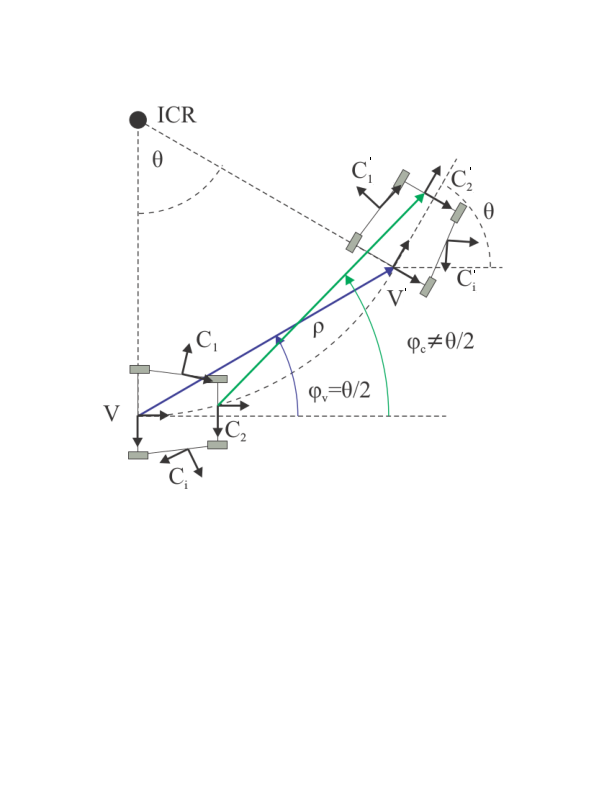}
    \caption{Relation between generalized camera in Ackermann motion \cite{leeCVPR13}.}
\label{img:Generalized_Circular}
\end{figure}

In order to reduce both the number of iterations and the number of solutions generated, we exploit the fact that 
a car typically follows the Ackerman motion constraint \cite{Siegwart11}, i.e., the car undergoes a circular motion on a plane around the Instantaneous Center of Rotation (ICR) (cf. \figref{img:Generalized_Circular}). In this case, the relative motion $(R,t)$ between consecutive frames can be parameterized with 2 degrees-of-freedom (DOF) as 
\begin{equation}
  R = \begin{bmatrix} \cos{\theta} & -\sin{\theta} & 0 \\ \sin{\theta} & \cos{\theta} & 0 \\ 0 & 0 & 1 \end{bmatrix},
  ~~t = \rho\begin{bmatrix}\cos{\varphi_{v}} \\ \sin{\varphi_{v}} \\ 0 \end{bmatrix} \enspace ,
  \label{eq:AckermannMotion}
\end{equation}
 where $\theta$ is the relative yaw angle and $\rho$ is the scale of the relative translation.
It can be easily 
seen from \figref{img:Generalized_Circular} that $\varphi_{v} = \frac{\theta}{2}$. Inserting \eqref{eq:AckermannMotion} into \eqref{eq:gec} and using two correspondences, we obtain a system of two polynomials (see our original publication \cite{leeCVPR13} for details) in the two unknowns $\rho$ and $\theta$. This system can be solved in closed form by finding the roots of a univariate cubic polynomial, resulting in up to 6 solutions.
Since only 2 correspondences are required, RANSAC only needs 16 iterations for $v=0.5$. 
Thus, applying our solver inside RANSAC is more efficient than using the 17-point and 6-point solvers, leading to an approach that is, in contrast to \cite{Li08,Pless03,Stewenius05}, applicable in practice.

\begin{figure}[t]
  \centering
      \includegraphics[width=1\columnwidth]{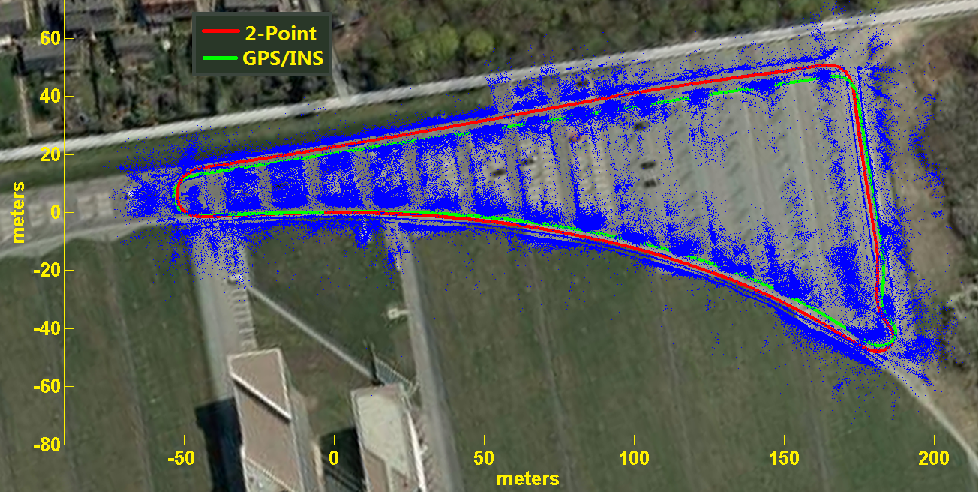}
    \caption{Top view of the trajectory and 3D map points estimated using our 2-point solver compared with GPS/INS ground truth \cite{leeCVPR13}.}
\label{img:overlaidmap}
\end{figure}

\figref{img:overlaidmap} shows a trajectory (red)
and 3D map points (blue) obtained by using our solver inside a SfM framework with bundle adjustment and loop-closure detection. 
As can be seen, the trajectory estimated from our 2-point algorithm is locally very close to the GPS/INS ground truth (green).  \textcolor{black}{Globally, the error is below 3m}. Some global shape difference is due to the fact that the mapped parking lot is not exactly planar.

\subsection{Loop-Closure Detection \& Handling}
\label{sec:sparse_mapping:loop_closure}
Small errors in the relative motion estimates eventually accumulate and lead to drift, i.e., a significant error in position and orientation when revisiting a place. Consequently, we need to perform loop-closure detection to identify previously visited locations, derive pose constraints between the current and previous visits, and optimize the trajectory and map accordingly.  
To this end, we form
a pose-graph \cite{IROS13_Lee_Robust}
$\mathcal{G} \in \{\mathcal{V}, \mathcal{E}\}$. 
The nodes $\mathcal{V} = {v_1, v_2,...,v_K}$ represent the global poses of the car, computed by concatenating  the relative poses estimated by our 2-point algorithm. 
\textcolor{black}{Among the} edges $\mathcal{E} = \{e_{1,2},e_{2,3},...,e_{i,j}\}$,
\textcolor{black}{an edge} $e_{i,i+1}$ thus corresponds to the
 relative motion estimated by the 2-point algorithm \textcolor{black}{between two subsequent time steps. In contrast, an edge} $e_{i,j}$, $j\neq i+1$, corresponds to \textcolor{black}{a pose estimated from non-sequential frames related via a} loop-closure \textcolor{black}{event}. 

Potential loop-closure edges are detected by vocabulary tree-based image retrieval \cite{Nister06}, 
followed by geometric verification \cite{LeeICRA14} to estimate the relative pose between frames. An edge $e_{i,j}$, $i\not=j$, is added to the pose graph if its inlier ratio is above a certain threshold \cite{LeeICRA14}. 
Given the new edges found during loop-closure detection, the pose graph is optimized as 
\begin{equation}\label{eq:posegraphOptimization}
\underset{v \in \mathcal{V}}{\operatorname{argmin}}\sum_{e_{i,j} \in\mathcal{E}}||h(v_{i}, v_{j}) - e_{i,j}||^2 \enspace .
\end{equation}
Here, $h(v_i, v_j)$, $i\not= j$, denotes the relative pose obtained from the two global poses $v_i$ and $v_j$. \eqref{eq:posegraphOptimization} thus aims to find a set of global positions that are as consistent as possible to the estimated relative motions. 
 We use the robust approach from our original publication \cite{IROS13_Lee_Robust} for pose graph optimization. 

The Ackermann motion model is only valid for two consecutive frames and is violated if the car revisits a place.  
Thus, our 2-point solver cannot be used for estimating the relative pose between loop closures. 
Instead, we only assume planar motion for a loop-closure candidate. Consequently, the relative pose $(R,t)$ has 3 DOF and is given by
\begin{equation}
  R = \frac{1}{1+q^{2}}\begin{bmatrix} 1-q^{2} & -2q & 0 \\ 2q & 1-q^{2} & 0 \\ 0 & 0 & 1+q^{2} \end{bmatrix},
  ~~t = \begin{bmatrix}x \\ y \\ 0 \end{bmatrix},
  \label{eq:planarMotion}
\end{equation}
where $q = tan(\frac{\theta}{2})$. 
Inserting \eqref{eq:planarMotion} into 
\eqref{eq:gec} and using 3 correspondences, we obtain a system of three polynomials (see our original publication \cite{IROS13_Lee_Structureless} for details) in the three unknowns $x$, $y$ and $q$.
Solving this system \cite{Cox97} results in up to six real solutions. Since only 3 correspondences are required, 
we only need $34$ RANSAC iterations for $v=0.5$.

Note that in case the planar motion assumption is violated, e.g., when the car is moving up a ramp, we can make use of the roll and pitch angles from an inertia measurement unit (IMU)
to efficiently estimate the motion with a 4-point algorithm as detailed in our previous work \cite{Lee14CVPR}.

\begin{figure}[t]
  \centering
\begin{subfigure}[b]{0.49\columnwidth}\centering
\includegraphics[width=1\columnwidth, trim = 20mm 15mm 30mm 15mm, clip]{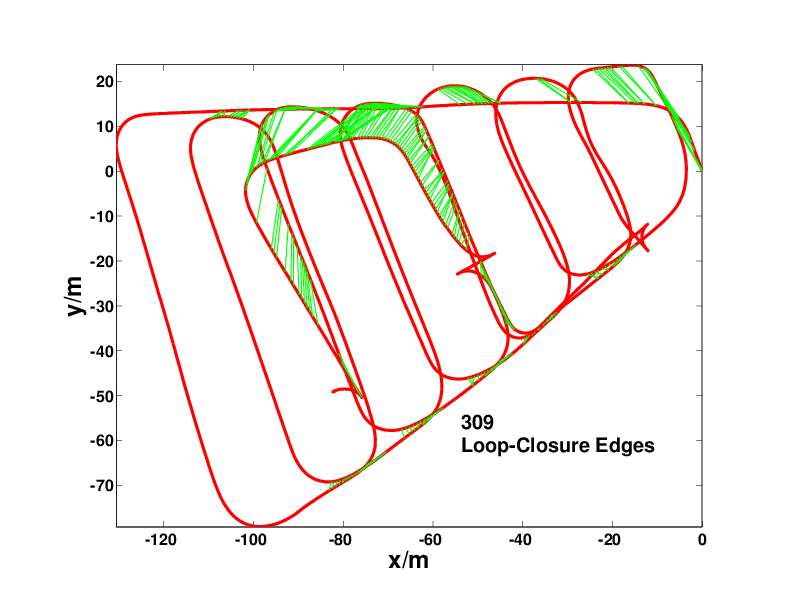}
\label{img:SuperMarket_ParkingGarage_beforeLoopClosure_generalized_beforeLoopClosure_monocularFront}
  \end{subfigure}
  \begin{subfigure}[b]{0.49\columnwidth}\centering
 \includegraphics[width=1\columnwidth, trim = 20mm 15mm 30mm 15mm, clip]{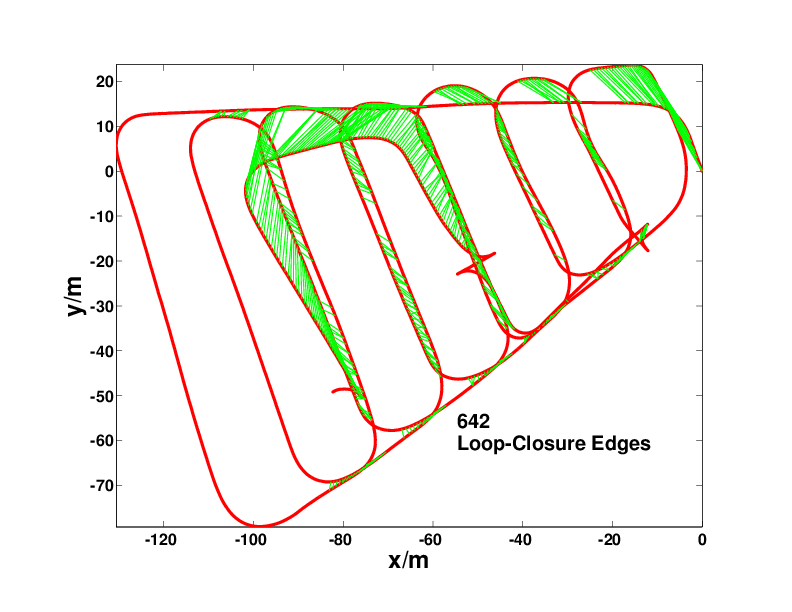}
\label{img:SuperMarket_ParkingGarage_beforeLoopClosure_generalized}
  \end{subfigure}
\caption{Pose-graph of the Parking Garage dataset from wheel odometry (red) before pose-graph optimization \cite{IROS13_Lee_Structureless}. 
Comparison on the total loop-closure edges (green) found from (left) a forward looking monocular camera and (right) our multi-camera system.}
\vspace{-0.5cm}
\label{img:SuperMarket_ParkingGarage_beforeLoopClosure}
\end{figure}

\figref{img:SuperMarket_ParkingGarage_beforeLoopClosure} shows a comparison of the total number of loop-closures detected by the monocular front-looking camera and multi-camera system. 
We can see that the multi-camera system leads to many more loop-closure detections especially in the parts where the car is moving in opposite direction to the previously visited path.

\begin{figure}[t]
  \centering
  \begin{subfigure}[b]{0.49\columnwidth}\centering
    \includegraphics[width=0.9\columnwidth, trim = 20mm 15mm 30mm 15mm, clip]{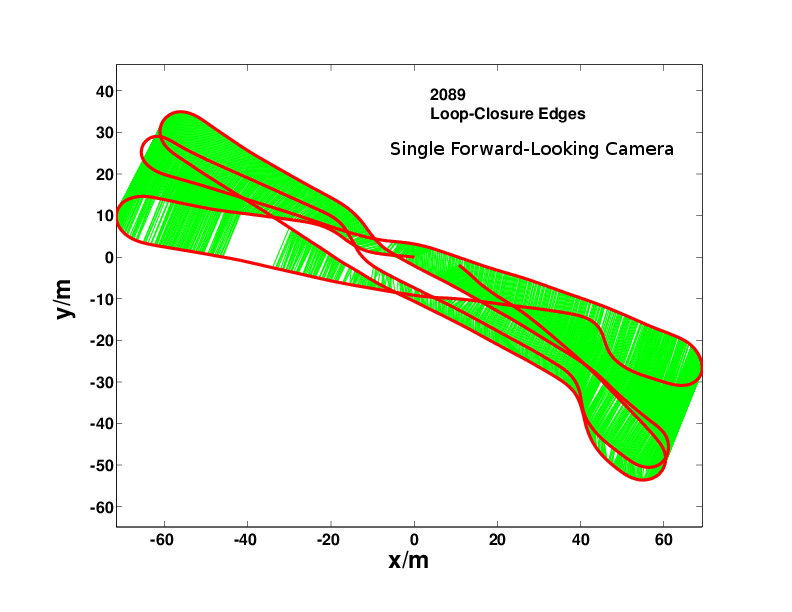}
    \label{img:hoenggerberg_monoFront}   
  \end{subfigure}
  \begin{subfigure}[b]{0.49\columnwidth}\centering
    \includegraphics[width=0.9\columnwidth, trim = 20mm 15mm 30mm 15mm, clip]{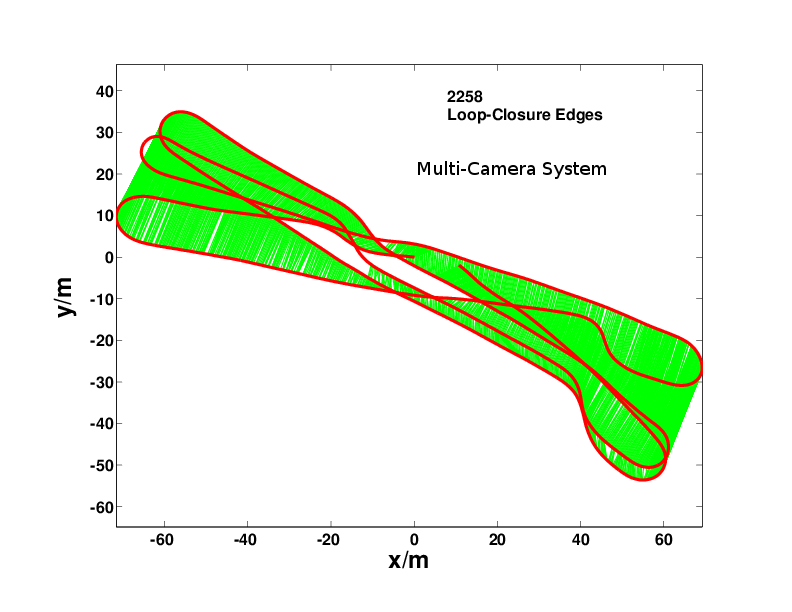}
    \label{img:hoenggerberg_generalized}
  \end{subfigure}
  \begin{subfigure}[b]{1\columnwidth}\centering
    \includegraphics[width=0.9\columnwidth]{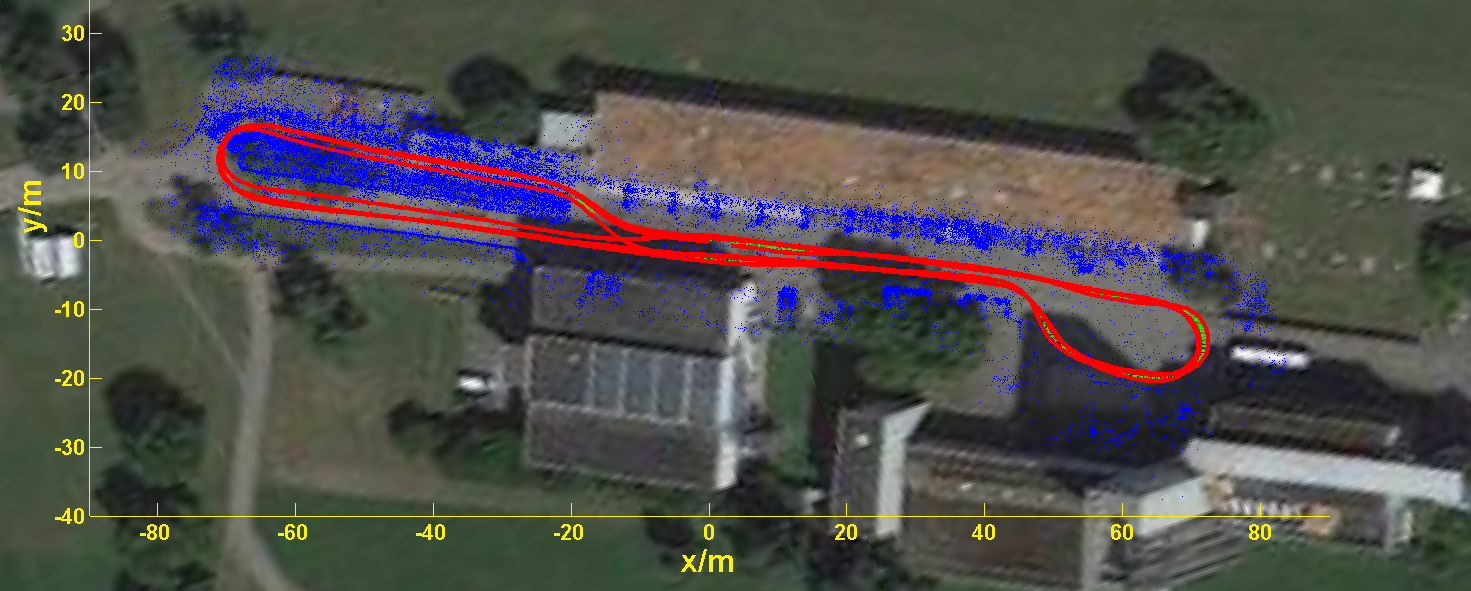}
    \label{img:hoenggerberg_Results}
  \end{subfigure}
  \caption{(Top row) Pose-graph of the Campus dataset from visual odometry (red) before pose-graph optimization \cite{IROS13_Lee_Structureless}. 
           Green lines denote loop-closure constraints found with (top left) a single forward-looking camera and (top right) our multi-camera system.
           (Bottom) Pose-graph (red) and 3D scene points (blue) for Campus dataset after pose-graph optimization overlaid on the satellite image.}
  \label{img:hoenggerberg}
\end{figure}

\figref{img:hoenggerberg} shows another example of the loop-closure detection before (top) and after (bottom) pose-graph optimization. 
We observe almost similar number of loop-closure detections for both the front-looking camera and multi-camera system in this example.  
This is because most of the loop-closure detection paths are facing the same directions. 
As can be seen, the reconstructed estimated 3D points after pose-graph optimization align well with the buildings shown in the satellite image.

\subsection{Summary}
More details of the minimal solvers for ego-motion estimation and loop-closure handling
can be found in our original publications \cite{leeCVPR13, IROS13_Lee_Structureless, Lee14CVPR}. Our minimal solvers forms the backbone of our sparse mapping framework. They enable us to 
map out large areas efficiently by treating multiple cameras that are fixed rigidly onto the car with know extrinsics parameters jointly as one. 

\begin{figure}[t]
  \centering
      \includegraphics[width=0.7\columnwidth, trim = 20mm 15mm 20mm 15mm, clip]{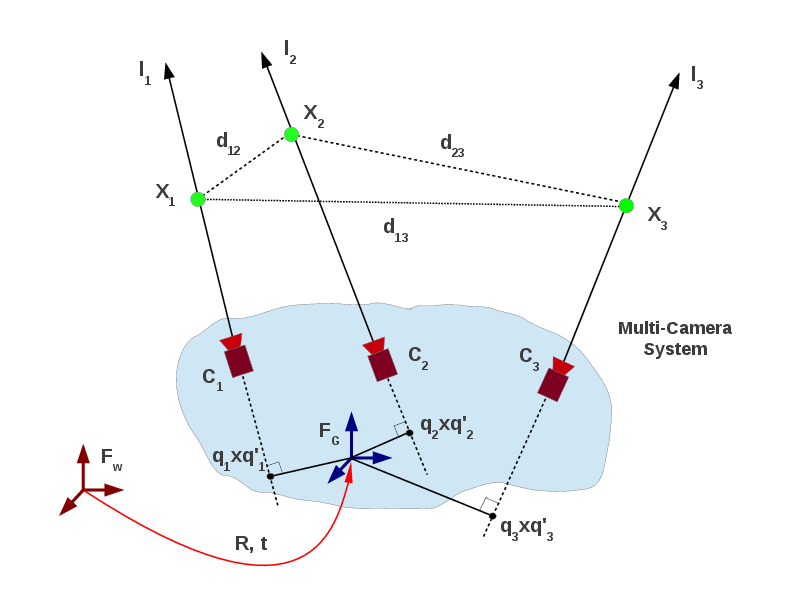}
    \caption{Illustration of the pose estimation problem for a multi-camera system \cite{leeISRR13,leeIJRR15}.}
\label{img:poseEstimationGeneralized}
\end{figure}

\section{Localization}
\label{sec:Localization}
Each 3D point in the sparse maps build with the approach from \secref{sec:sparse_mapping} has been triangulated from multiple 2D image features, e.g., SURF features \cite{Bay2008}. Thus, the maps can also be used for online localization of the car: Given the features extracted from novel images, 2D-3D matches between the image features and the 3D points in the map can be established by comparing the associated feature descriptors \cite{sattler11}. The global pose of the cameras wrt. the map can then be estimated from these 2D-3D correspondences.

Formally, localization refers to the problem of determining the rigid 
transformation $(R,t)$ between the fixed world frame $F_W$ of the map, and the multi-camera frame $F_G$, given a set of 2D-3D correspondences (cf. \figref{img:poseEstimationGeneralized}).
Obviously, the pose could be estimated for each camera individually using standard pose solvers 
\cite{haralick1991pose, Long99, Moreno07}. However, this approach would ignore that we know the geometric relation between the cameras in the multi-camera system. In addition, using a larger FOV usually leads to more accurate pose estimates. 
We thus derived an approach for estimating the global pose of a generalized camera  
\cite{leeISRR13,leeIJRR15}, which we outline below.
As is the case for calibrated perspective cameras \cite{haralick1991pose}, our solver requires three 2D-3D correspondences for global pose estimation.

As for the relative pose solvers presented above, we again use Pl\"{u}cker lines to handle the fact that not all light rays meet in a single center of projection. 
The 3D position of a point $X_{i}^{G}$ on a Pl\"{u}cker line defined in $F_G$ is given by 
\begin{equation}
  X_{i}^{G} = q_{i} \times q_{i}' + \lambda_{i}q_{i} \enspace,
  \label{eq:plueckerLine}
\end{equation}
where $\lambda_{i}>0$ is the depth of the point.
As for perspective cameras \cite{haralick1991pose}, the distances $d_{ij}$ (cf. \figref{img:poseEstimationGeneralized}) between the 3D points $X_{i}$ in  $F_W$ and the points $X_{i}^{G}$ in  $F_G$ need to be the same. 
Using this constraint and the Pl\"{u}cker line representation, we can derive three equations that can be used to solve for the three unknown depths $\lambda_i$. 
Knowing the depths defines the 3D point positions $X_{i}$ in $F_G$ and we can compute the global pose $(R,t)$ that aligns the $X^G_i$ and the $X_{i}$ using absolute orientation \cite{haralick1991pose}.

\begin{figure}[h]
  \centering
  \begin{subfigure}[b]{0.49\columnwidth}\centering
  \includegraphics[width=1\columnwidth, trim = 0mm 0mm 0mm 0mm, clip]{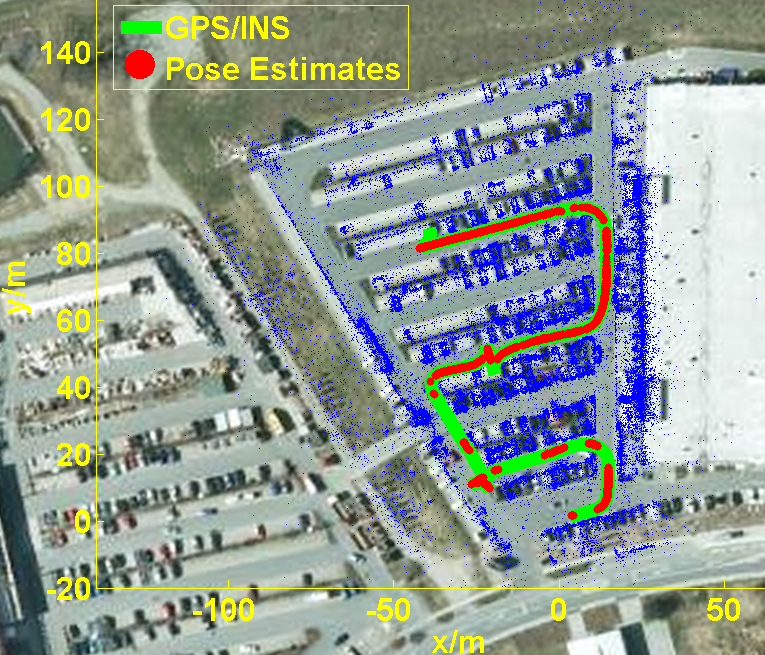}
  \label{img:SuperMarket_results}
  \end{subfigure}
  \begin{subfigure}[b]{0.49\columnwidth}\centering
  \includegraphics[width=1\columnwidth, trim = 30mm 10mm 30mm 10mm, clip]{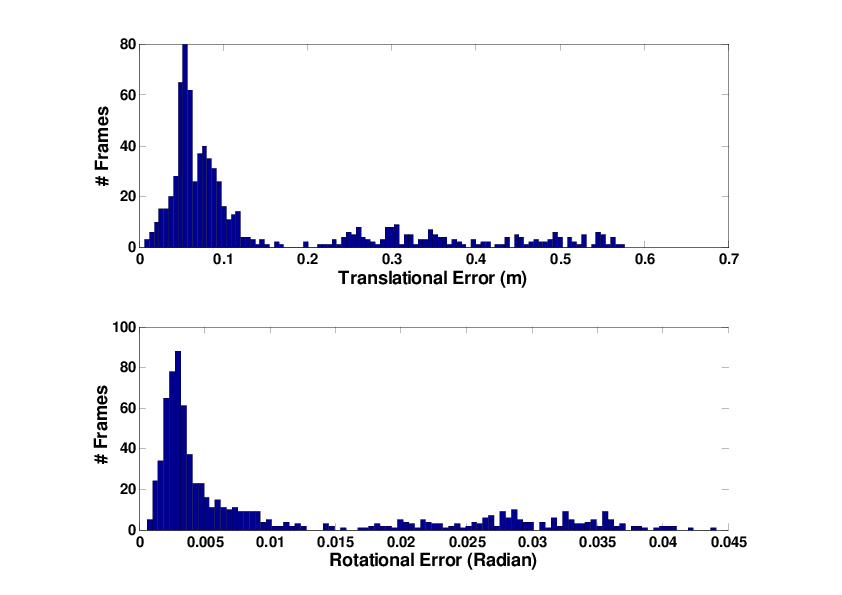}
  \label{img:SuperMarket_matches_errors}
  \end{subfigure}
  \caption{(a) Localization results for the Supermarket dataset. Results from frames with $<20$ correspondences are discarded.  
  (b) Plots showing the distribution of the translational and rotational errors against GPS/INS ground truths \cite{leeISRR13,leeIJRR15}.}
  \label{img:SuperMarket}
\end{figure}

\figref{img:SuperMarket} shows the results of our visual localization algorithm. We first 
collected a set of images with GPS/INS ground truth to build the sparse map (blue points). We collect a separate set of images with GPS/INS ground truth (green trajectory) 
at another time for testing our localization algorithm. The poses estimated with our algorithm are shown in red. As can be seen, our algorithm produces accurate localization results with only small errors. Note that the missing parts of the trajectory is due to failure in visual recognition because of changes in the scene. Such failures are typically not a problem in practice as these missing parts can be bridged via the car's wheel odometry. More details can be found in our previous works \cite{leeISRR13,leeIJRR15}. 
\section{Dense Mapping}
\label{sec:denseMapping}

Surfaces such as building facades and the ground are generally not captured well in the sparse maps.
This section explains how dense multi-view stereo matching can be done directly on the fisheye images and how the depth maps can be fused into a dense height map representation of the environment. The output of the multi-view stereo matching are depth maps which can be computed in real-time and can be directly used during fully automated driving missions.
The dense height map is computed in an offline manner and is used to map an environment for later automated driving within this space.

\subsection{Depth Map Computation}

We use plane sweeping  stereo matching \cite{collins1996space,yang2003multi} which has the strength that multiple images can be matched without prior stereo rectification. Meaning, the epipolar lines do not need to be aligned with the image scan lines. Knowledge about predominant surface directions can be incorporated \cite{gallup2007real} and we extended the algorithm to directly match fish eye images \cite{hane2014real}.

Images from a similar viewpoint are compared to a reference image by projecting the images onto a plane which serves as proxy geometry. If the plane coincides with the original geometry the projection of the images has high similarity. Instead of projecting to the plane, the reference image is fixed and the other images are warped to reference image via the proxy plane, which is a planar homography for pinhole images. For fish eye images the camera model can be included. The comparison of the images takes place in the reference view. Choosing the plane with lowest dissimilarity for each pixel leads to a dense depth map for the reference image.

We follow our original paper \cite{hane2014real} and refer the reader for more details. Given is a set of images $\mathcal{I} = \{I_1,\ldots,I_n,\ldots,I_N\}$ (tangential and radial distortion removed) and the respective camera parameters $\mathcal{P} = \{P_1,\ldots.P_n,\ldots,P_N\}$, with $P_n = \{\phi_n, \mathbf{R}_n, \mathbf{C}_n\}$. The camera intrinsics $\phi = \{\xi, \mathbf{K}\}$ are composed of the parameter $\xi$ of the unified projection model \secref{sec:cameraModel} and the intrinsic camera projection matrix $\mathbf{K}$. The projection of a 3D point $\mathbf{X}$ to the image plane is described as $[\mathbf{x}^T, 1]^T  = \mathbf{K}\hbar(\xi, \mathbf{R}\mathbf{X} - \mathbf{R}^T\mathbf{C})$. The inverse operation maps an image pixel $\mathbf{x}$ to a ray from the camera center $\mathbf{C}$ to a point on the ray defined as $\mathbf{X} = \mathbf{R}\hbar^{-1}(\xi, \mathbf{K}^{-1}[\mathbf{x}^T, 1]^T) + \mathbf{C}$. Finally, we need a set of planes $\Pi = \{\pmb{\Pi}_1,\ldots,\pmb{\Pi}_m,\ldots,\pmb{\Pi}_M\}$, which serve as proxy geometry to test the image dissimilarity at different depths. They are defined as $\pmb{\Pi} = [ \mathbf{n}^T, d]$, with $\mathbf{n}$ being the unit-length normal direction pointing to the camera and $d$ the distance of the plane to the camera center. The transformation which warps image $I_n$ to the reference image $I_\mathrm{ref}$ using plane hypothesis $\Pi_m$ is then given as,
\begin{align}
 \mathbf{H}_{n,m}^{\mathrm{ref}} &= \mathbf{R}_n^T\mathbf{R}_{\mathrm{ref}} + \frac{1}{d_m}(\mathbf{R}_n^T \mathbf{C}_n - \mathbf{R}_n^T\mathbf{C}_{\mathrm{ref}})\mathbf{n}_m^T  \nonumber \\
 [\mathbf{x}^T_n, 1]^T & = \mathbf{K}_n \hbar(\xi_n, \mathbf{H}_{n,m}^{\mathrm{ref}} \hbar^{-1}(\xi_{\mathrm{ref}}, \mathbf{K}_{\mathrm{ref}}^{-1} [\mathbf{x}^T_{\mathrm{ref}}, 1]^T)). \label{eq:warping}
\end{align}

Each image $I_n$, with $n \neq \mathrm{ref}$ is warped into an image $I_{n,m}^{\mathrm{ref}}$ for each plane $\pmb{\Pi}_m$ using \eqref{eq:warping} and a pixel-wise image dissimilarity $\mathbf{D}_{n,m}^\mathrm{ref}$ is computed between each of the warped images and the reference image. We use the negative zero mean normalized cross correlation (ZNCC) as image dissimilarity measure. For each pixel and plane we have several image dissimilarity measures originating from all the images $I_n$. In order to have a single dissimilarity measure for each pixel and plane they are aggregated into a single $\mathbf{D}_m^\mathrm{ref}$. We mention two choices, the first one is simply averaging the costs. This is computationally the most efficient variant and is hence used for online applications. The second one is very suitable for images taken in sequence, such as images taken with a single camera while a car is moving. One average is computed over all the images which were taken before the reference image and one average over all of the images which are taken after the reference image. Finally, the one which has a lower dissimilarity is the final aggregated cost. The advantage of this procedure is that it is more robust to occlusions than simple averaging \cite{kang2001handling} but needs to know images taken in the future with respect to the reference image and is therefore used for offline mapping in our system.

The final depth map is extracted by choosing for each pixel the plane $\hat{m}$ with lowest dissimilarity measure and \textcolor{black}{computing} the induced depth \textcolor{black}{as}
\begin{equation}
 Z_{\hat{m}}(\mathbf{x}) = -\frac{d_{\hat{m}}}{\mathbf{n}_{\hat{m}}^T\hbar(\xi_{\mathrm{ref}}, \mathbf{K}^{-1}_{\mathrm{ref}} [\mathbf{x}^T,1]^T)} \enspace.
\end{equation}
To reduce the artifacts of the depth discretization we use sub-plane interpolation by fitting a parabola \cite{tian1986algorithms} into the scores of the best plane and the neighboring planes with the same normal direction. The depth is then chosen at the position where the parabola has its minimum.

\begin{figure}
\centering{
 \includegraphics[width=0.45\linewidth]{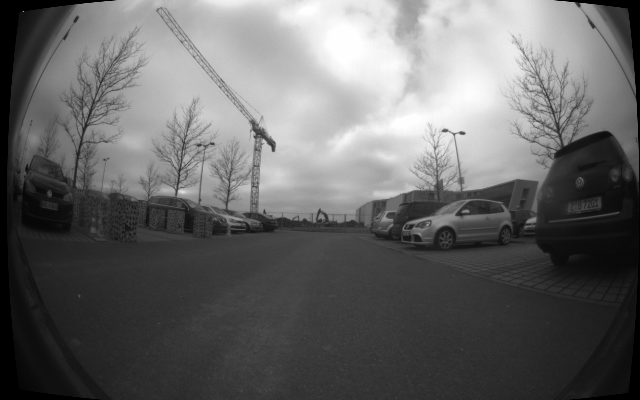}
 \includegraphics[width=0.45\linewidth]{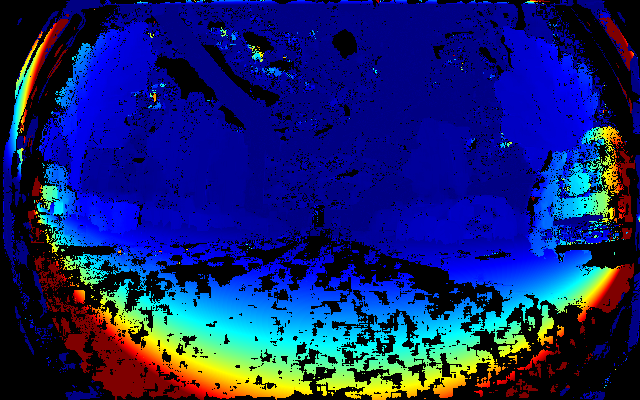} \vspace{0.1cm} \\
 \includegraphics[width=0.45\linewidth]{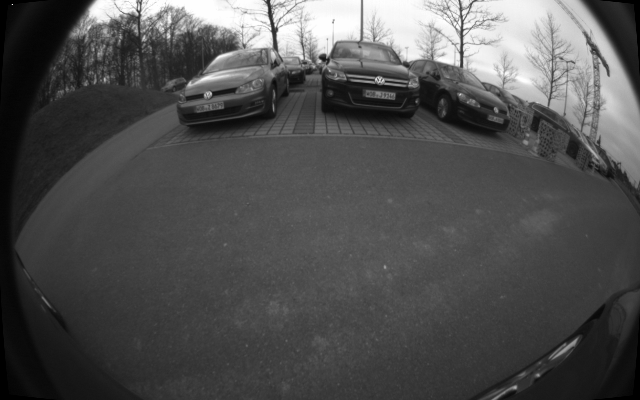}
 \includegraphics[width=0.45\linewidth]{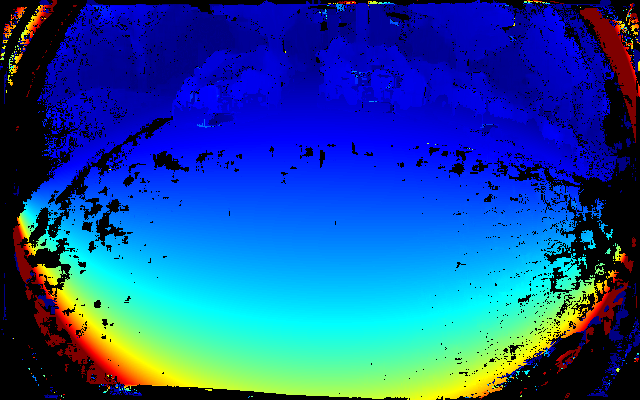}
 }
 \caption{Two example depth maps next to the corresponding reference image. The depth is color codeded with dark red corresponding to the selected near distance and dark blue to selected far distance. Note that the ground directly next to the front wheel in the lower image is correctly reconstructed.}
 \label{fig:exampleDepthMaps}
\end{figure}

The intrinsic camera parameters $\phi$ are determined during the camera calibration procedure \secref{sec:calibration}. The camera poses are either coming directly from the sparse map computed during the extrinsic calibration \secref{sec:slam_extrinsic_calibration} or during sparse mapping \secref{sec:sparse_mapping}. For online mapping the vehicle's wheel odometry poses can be used (cf.\ \secref{sec:obstacleDetection}). As input images we select a sequence of a few images captured by a single camera while the vehicle is moving. No matching between different cameras is done. In order to select the planes we first select the normal direction of the planes. The most important direction is fronto-parallel to the image plane. Choosing planes in this direction is very similar to the rectified two-view stereo matching using disparities along scan lines. A near and a far distance to the camera determines the region in which we want to measure the geometry. This is typically a range from 30cm to 50m. Within this range the planes are placed such that the difference of inverse distance $1/d$ to the neighbors is constant, which corresponds to disparities. Note that the sidewards facing cameras are pointing towards the ground and therefore a near distance of 30cm to the camera allows us to reconstruct the ground even right next to the car (cf. \figref{fig:exampleDepthMaps}). For a more faithful reconstruction of the ground, planes parallel to the ground can be used in addition. We use this for online obstacle detection \secref{sec:obstacleDetection}. \figref{fig:exampleDepthMaps} shows example depth maps acquired using our method.

\subsubsection{Summary}
In this section we presented a system which allows to directly match fisheye images. It has a less then 10\% running time increase compared to using images with the same resolution and a pinhole projection model. This allows us to get a full coverage of the field of view of the fisheye camera with very little runtime overhead over first unwarping to a pinhole projection. Further evaluations including run-time performance measurements can be found in \cite{hane2014real} and our stereo matching code is publicly available\footnote{\url{http://www.cvg.ethz.ch/research/planeSweepLib/}}.

\subsection{Height Map Fusion}
Our height map fusion approach is a modified version of our method \cite{hane2011stereo}, which we initially proposed for indoor environments. We give a brief overview of the method focusing on the modified steps and refer the reader to \cite{hane2011stereo} for more details. First, all the depth maps are fused into a voxel grid and afterwards a regularized height map is extracted out of the voxel grid. For each voxel $(x,y,z)$ a signed scalar weight $w_{(x,y,z)}$ is stored which indicates how strongly the data indicates that this voxel lies in the free (negative weight) or occupied space (positive weight). This weights are computed from the input depth maps.

For the next steps it is important to mention that the $z$-axis of the voxel grid is aligned with the vertical direction. In order to compute the final height map a set of values are extracted for each $z$-column of the voxel grid on which a $2$-pass optimization is executed. The first pass of the optimization divides the domain $\Omega$ of the height map into a region which is observed $\Omega_{\mathrm{obs}}$ and a non-observed region. The second pass determines the height values for locations within $\Omega_{\mathrm{obs}}$. Note that once we have the relevant information extracted from each $z$-column it can be discarded and hence the entire voxel grid never needs to be stored. First, we extract for each point $(x,y)$ an optimal height based on the data of just a single $z$-column \cite{gallup2010heightmap} by minimizing the cost function
\begin{equation}
 C_{(x,y)}(h) = - \sum_{z < h} w_{(x,y,z)} + \sum_{z \geq h} w_{(x,y,z)} \enspace. \label{eq:perColumnMin}
\end{equation}

For the first pass of the optimization we need two values, $C_{(x,y)}^{\mathrm{min}}$ which defines the minimal cost and $C_{(x,y)}^{\mathrm{occ}}$, which defines the cost if we assume that the entire column is in the occupied space. The rationale behind this is that we only want to compute a height value if we are reasonably sure that we did observe free space at a specific location. In order to avoid having a noisy region $\Omega_{\mathrm{obs}}$ we penalize its boundary length by minimizing the convex energy functional
\begin{align}
 E^{\mathrm{Labeling}}(l) = &\int_{\Omega} \| \nabla l(\mathbf{x})  \|_2 + \lambda_l (l(\mathbf{x})(C_{(x,y)}^{\mathrm{min}} +  \gamma) + (1-l(\mathbf{x}))C_{(x,y)}^{\mathrm{occ}}) \mathrm{d}\mathbf{x} \\
 & \text{subject to } l(\mathbf{x}) \in [0,1].
\end{align}
The function $l(\mathbf{x})$ indicates if a location is included in $\Omega_{\mathrm{obs}}$. The parameter $\gamma$ is a penalty for including a location into the observed space and is used to adjust how strongly the free space needs to be observed. The parameter $\lambda_l$ adjusts the influence of the total variation (TV) smoothness prior which penalizes the boundary length. The functional is minimized using the first order primal-dual algorithm \cite{chambolle2011first}.

The second pass of the optimization is again posed as convex optimization. The data is represented as a convex approximation of \eqref{eq:perColumnMin} locally around the per-location minimal cost height $\hat{h}_{(x,y)}$ with a function of the form
\begin{equation}
 C_{(x,y)}^{\mathrm{conv}}(h) = \alpha_1 [\hat{h}_{(x,y)} - h - \delta]_{+} + \alpha_2 [h - \delta - \hat{h}_{(x,y)}]_{+} \enspace,
\end{equation}
with $[\cdot]_{+} = \max\{0,\cdot\}$. The parameters $\alpha_1$ and $\alpha_2$ steer how strongly the cost increases while we move away from the locally best position. We propose to use a parameter $\delta$ which ensures that the function is flat around the minimum and models the uncertainty due to the discretization and is hence chosen as half the discretization step. In the final optimization pass we propose to regularize the height map with a Huber total variation smoothness prior
\begin{equation}
 E^{\mathrm{Height}} = \int_{\Omega_\mathrm{obs}} \| \nabla h(\mathbf{x}) \|_{\varepsilon} + \lambda_h C_{(x,y)}^{\mathrm{conv}}(h(\mathbf {x}))\mathrm{d}\mathbf{x} \enspace.
\end{equation}
$\|\cdot\|_{\varepsilon}$ denotes the Huber norm which is defined as
\begin{equation}
 \|\mathbf{x}\|_\varepsilon = \begin{cases} \|\mathbf{x}\|^2_2/(2\varepsilon) & \text{if }  \|\mathbf{x}\|_2 \leq \varepsilon \\
 \|\mathbf{x}\|_2 - \varepsilon/2 & \text{if }  \|\mathbf{x}\|_2 > \varepsilon.\end{cases}
\end{equation}
Using the Huber norm leads to smooth height map and allows for discontinuities. The first order primal-dual algorithm \cite{chambolle2011first} is used for numerical minimization.

\begin{figure}
\centering{
 \includegraphics[width=0.75\linewidth]{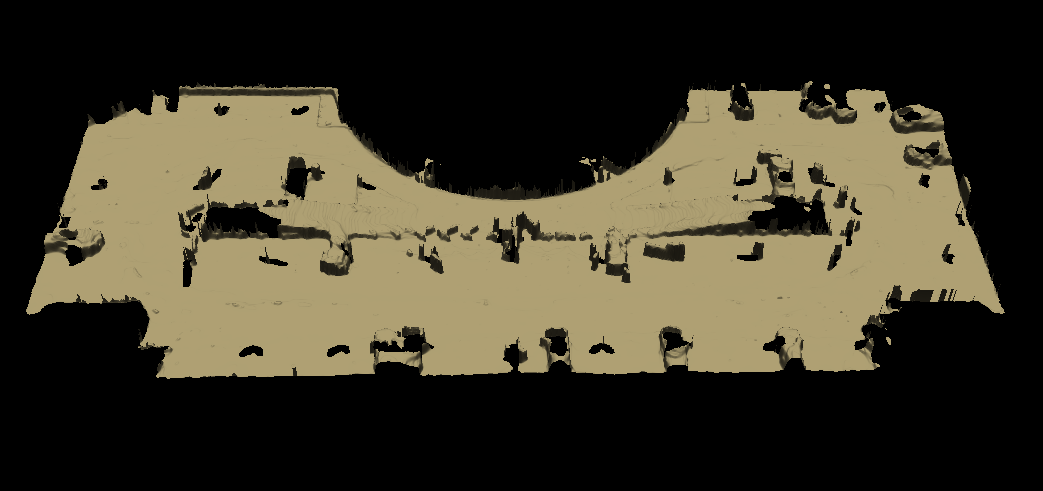} \\
 \includegraphics[width=0.75\linewidth]{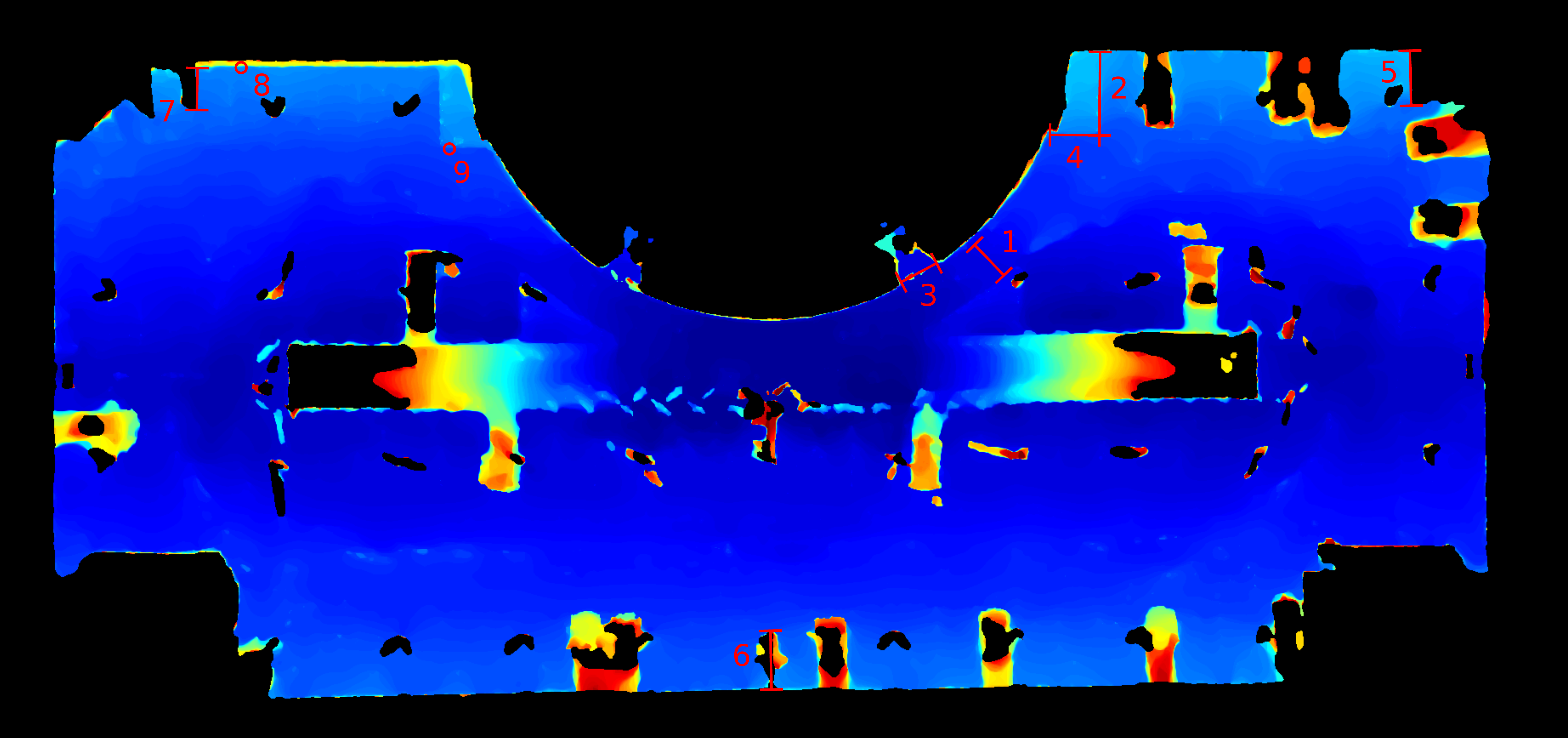} \includegraphics[height=5cm]{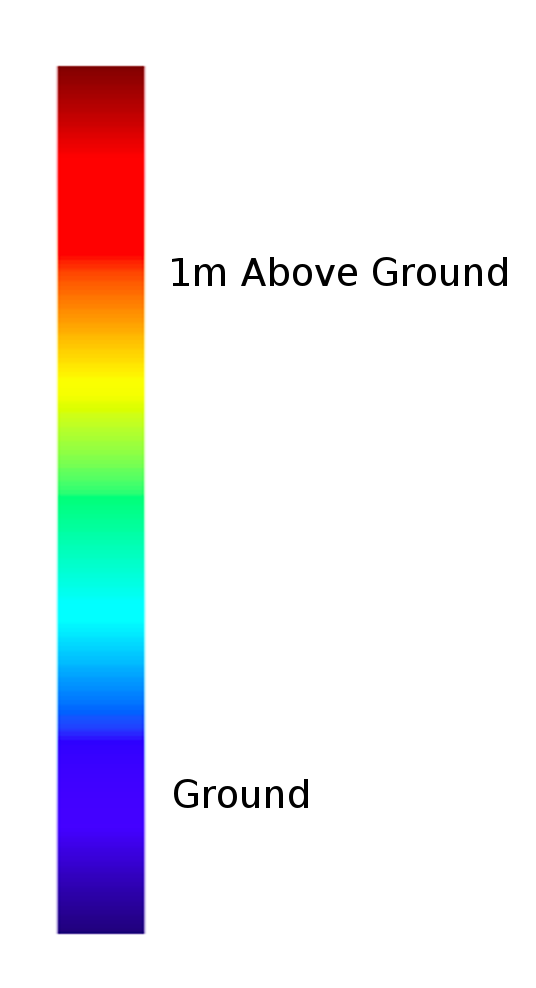}}
 \caption{(Top) Height map visualized as mesh. (Bottom) color coded height map (dark blue lowest, dark red highest) with the positions of the $9$ measurements indicated in red. Measurements $1$ to $7$ are length measurements and measurements $8$ and $9$ measure the height difference that occurs in the indicated circle.}
 \label{fig:heightQuantAnalysis}
\end{figure}

\begin{figure*}[thpb]
   \centering
   \includegraphics[width=\textwidth]{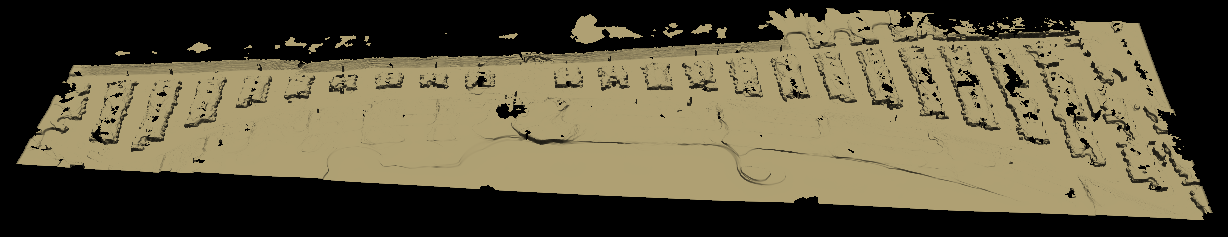}
   \caption{Height map of a large car park visualized as mesh.}
   \label{fig:dense_map}
 \end{figure*}

In order to quantitatively asses the performance of our method we mapped an underground parking garage and afterwards measured various horizontal and vertical distances using a tape measure as ground truth. The quantitative results presented in \figref{fig:heightQuantAnalysis} and \tabref{tab:heighQuantAnalysis} show that the map is accurate enough for autonomous driving with a mean error of $7.06$cm over 9 measurements. An example of a larger car park is depicted in \figref{fig:dense_map}.

\subsubsection{Summary}
In this section we presented a method which fuses depth maps into a height map representation. Our quantitative evaluation shows that the error is usually within a few centimeters. Our method allows us to map large parking spaces and efficiently store them as height map. Also refer to an earlier version of our algorithm in \cite{hane2011stereo}.

 \begin{table}
 \centering
 \footnotesize{
 \begin{tabular}{|l|c|c|c|r|}
 \hline
 Nr. & Ground truth & Map & Error \\
 \hline \hline
 $1$ & $2.725$m & $2.650$m & $7.5$cm \\
 \hline
 $2$ & $4.895$m & $4.800$m & $9.5$cm \\
 \hline
 $3$ & $2.570$m & $2.485$m & $8.5$cm \\
 \hline
 $4$ & $2.740$m & $2.900$m & $16.0$cm \\
 \hline
 $5$ & $3.140$m & $3.100$m & $4.0$cm \\
 \hline
 $6$ & $3.450$m & $3.400$m & $5.0$cm \\
 \hline
 $7$ & $2.505$m & $2.450$m & $5.5$cm \\
 \hline
 $8$ & $0.850$m & $0.800$m & $5.0$cm \\
 \hline
 $9$ & $0.125$m & $0.100$m & $2.5$cm \\
 \hline
 \end{tabular}
 }
 \caption{Quantitative evaluation for the $9$ indicated measurements in \figref{fig:heightQuantAnalysis}. The mean error over all the $9$ measurements is $7.06$cm.}
 \label{tab:heighQuantAnalysis}
 \end{table}

\section{Applications}
\label{sec:applications}

\subsection{Structure-based Calibration}

Both the SLAM-based extrinsic calibration and sparse mapping pipelines described in \secref{sec:slam_extrinsic_calibration} and \secref{sec:sparse_mapping} respectively generate a sparse map of the environment. This sparse map is used for visual localization in \secref{sec:Localization}. This map can also be used for structure-based calibration \citep{Heng2014ICRA,Heng2015JFR} which leverages visual localization to calibrate the multi-camera system. Here, we localize each camera individually as the extrinsic parameters are unknown. In contrast, in \secref{sec:Localization}, we localize a single multi-camera system assuming that the extrinsic parameters are known. By using natural features instead of fiducial targets, we minimize infrastructure setup costs.

\begin{figure}[t]
\centering
\includegraphics[width=0.8\textwidth]{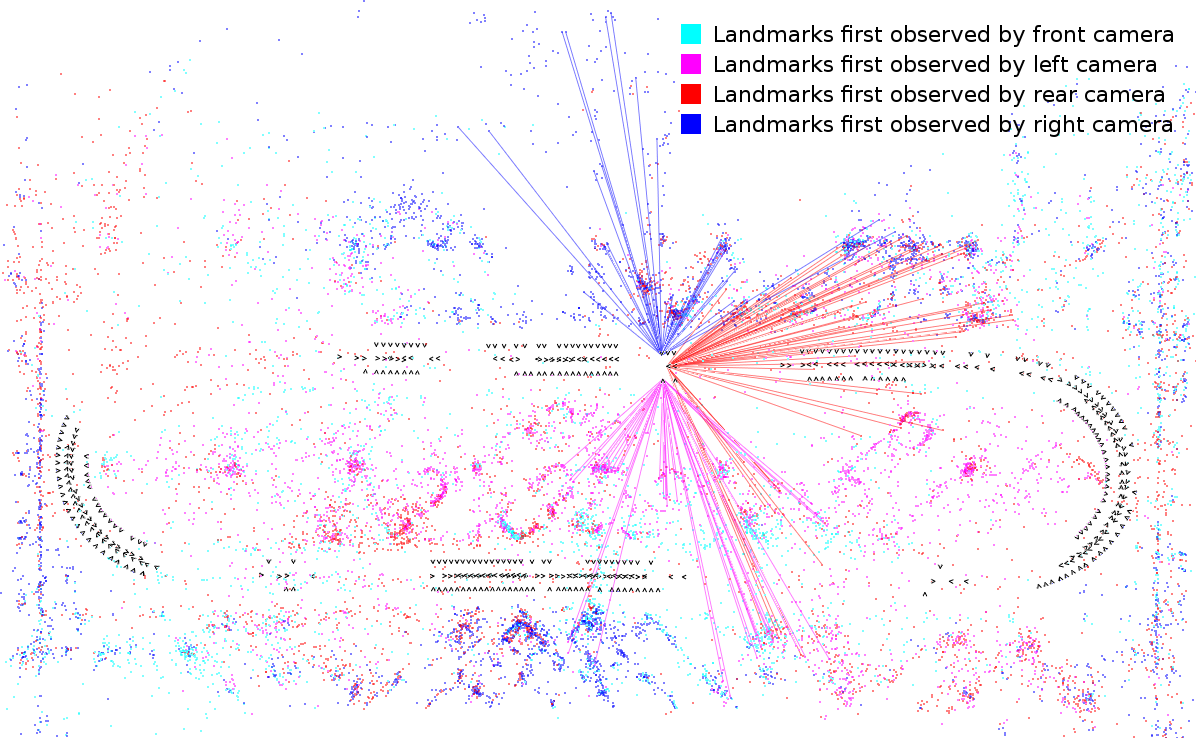}
\caption{Points represent 3D map landmarks which are colored according to the camera they were first observed in. Small black triangles correspond to estimated camera poses. Lines connect currently observed landmarks to the current camera poses.}
\label{fig:structure_calibration}
\end{figure}

With a sparse map of the calibration area, we use visual localization over a frame sequence to obtain 2D-3D feature correspondences for each camera, and estimate the camera poses. In turn, we infer an initial estimate of the extrinsic parameters and vehicle poses, and optimize these variables by minimizing the sum of weighted squared reprojection errors over all 2D-3D feature correspondences. \figref{fig:structure_calibration} illustrates the visual localization process. This structure-based calibration method is computationally efficient and runs in near real-time compared to the SLAM-based calibration method which takes a few hours to process 500 frames from a four-camera system. Prior knowledge of the 3D landmark coordinates reduces the algorithmic complexity.

Results quantifying the accuracy and repeatability of our structure-based calibration can be found in \citep{Heng2014ICRA,Heng2015JFR}. We attribute the accuracy and repeatability of our structure-based calibration to two reasons: the sparse map of the calibration area is accurate, and we use many 2D-3D feature correspondences over many frames for the calibration. Our structure-based calibration allows us to calibrate many multi-camera systems with high accuracy, in a short time, and without the need for infrastructure.

\subsection{Obstacle Detection}
\label{sec:obstacleDetection}

The depth maps from \secref{sec:denseMapping} are also used during automated driving in form of an obstacle detection system which facilitates online obstacle extraction \cite{hane2015obstacle}. In order to run in real-time we reason about the obstacles in 2D. Everything sticking out of the ground plane, which is given by the car calibration, is considered an obstacle.

Our algorithm extracts obstacles from each depth map independently. First, a 2D occupancy grid is generated from which obstacle hypotheses are extracted along rays. For each ray the closest occupied cell to the car is considered as obstacle. The final position is subsequently refined taking into account all the 3D points which fell into this specific cell.

The obstacle extraction procedure filters a lot of the noise as it combines many 3D points from the depth map into a single obstacle. The remaining noise is removed by fusing obstacle maps over time and from multiple cameras into a consistent 2D obstacle map. Most of the computation is done on GPU. The whole obstacle detection system is running in real-time (12.5Hz capturing frequency) on the left, right and front facing camera on a single computer with an NVIDIA GeForce GTX 680 GPU. The obstacle map is accurate enough for low speed driving and parking in our experiments we measure errors of less than 10cm, see \cite{hane2015obstacle} for the full evaluation.

\begin{figure}
 \centering
 \includegraphics[height=1.75cm]{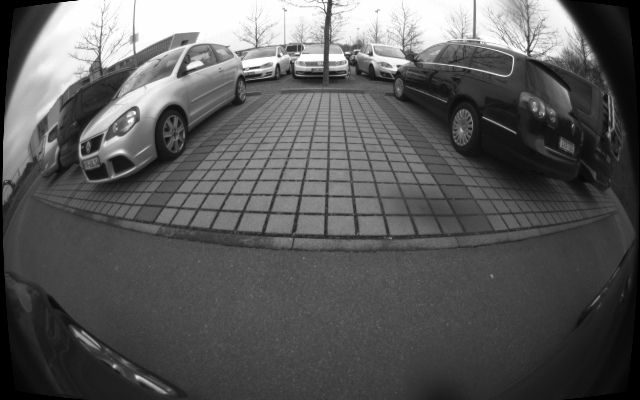}
 \includegraphics[height=1.75cm]{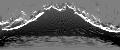}
 \includegraphics[height=1.75cm]{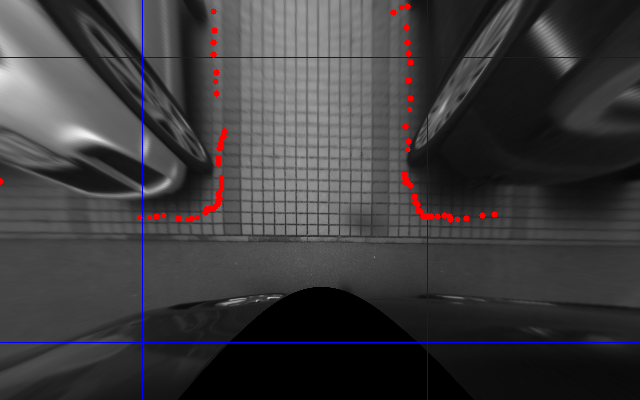}
 \includegraphics[height=1.75cm]{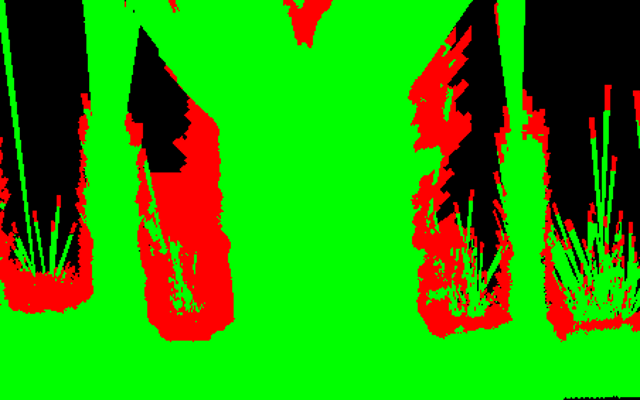}
 \caption{(left to right) reference image, occupancy grid (white occupied, black free), obstacles, fused map (green free, red occupied, black unobserved)}
\end{figure}

\section{Conclusions}
\textcolor{black}{In this paper, we have considered the problem of 3D perception from a multi-camera system in the context of self-driving cars. }
We \textcolor{black}{have} presented a pipeline which facilitates calibration of a multi-camera system composed of four fisheye cameras, computation of metric sparse and dense maps, map-based localization, and online obstacle detection. \textcolor{black}{This system has been used successfully on the self-driving cars of the V-Charge project, demonstrating the practical feasibility of our pipeline. Due to space constraints, this paper has focused } on a description of the pipeline as a whole and how the different components interact with each other. \textcolor{black}{For technical details and more thorough experimental results, we refer the interested reader to the original publications referenced in each section.} 

The chosen camera system contains four fisheye cameras which are mounted on the car such that they have minimal overlap allowing a 360 degree coverage. While more elaborate systems with binocular pairs would be possible, our system only uses a small number of cameras to keep cost and complexity low. Using more cameras would lead to a denser sensing of the environment but would also increase complexity and cost of the system. \textcolor{black}{Naturally, an interesting question is to exploit the trade-off between these factors. Unfortunately, this was not possible due to budget restrictions in the V-Charge project. Instead, a } system specifically designed to be (cost) minimal \textcolor{black}{was used}. We show that many of the task required for self driving car applications at low speeds can be implemented using such a camera setup. We are able to handle the strong distortion of the fisheye lenses by using an appropriate projection model. \textcolor{black}{By}  working directly on the fisheye images\textcolor{black}{, we were specifically able to avoid} any unwarping to a pinhole projection model \textcolor{black}{and thus were able to avoid a loss in the usable field-of-view}. 

Future work will address dynamic objects and semantic segmentation. For example, knowledge about the semantic class (e.g.\ tree, car or lane marking) of specific regions in the sparse or dense map could provide valuable information for localization and obstacle detection.

\section{Acknowledgments}

This work is supported in parts by the European Community's Seventh Framework Programme (FP7/2007-2013) under grant \#269916 (V-Charge) and 4DVideo ERC Starting Grant Nr. 210806.



\section*{References}
\bibliographystyle{abbrvnat}
\bibliography{references}

\begin{thebibliography}{43}
\providecommand{\natexlab}[1]{#1}
\providecommand{\url}[1]{\texttt{#1}}
\expandafter\ifx\csname urlstyle\endcsname\relax
  \providecommand{\doi}[1]{doi: #1}\else
  \providecommand{\doi}{doi: \begingroup \urlstyle{rm}\Url}\fi

\bibitem[Bay et~al.(2008)Bay, Ess, Tuytelaars, and Van~Gool]{Bay2008}
H.~Bay, A.~Ess, T.~Tuytelaars, and L.~Van~Gool.
\newblock Speeded-up robust features (surf).
\newblock In \emph{Computer Vision and Image Understanding (CVIU)}, 2008.

\bibitem[Carrera et~al.(2011)Carrera, Angeli, and Davison]{carrera11}
G.~Carrera, A.~Angeli, and A.~Davison.
\newblock Slam-based automatic extrinsic calibration of a multi-camera rig.
\newblock In \emph{IEEE International Conference on Robotics and Automation
  (ICRA)}, pages 2652--2659, 2011.

\bibitem[Chambolle and Pock(2011)]{chambolle2011first}
A.~Chambolle and T.~Pock.
\newblock A first-order primal-dual algorithm for convex problems with
  applications to imaging.
\newblock \emph{Journal of Mathematical Imaging and Vision}, 40\penalty0
  (1):\penalty0 120--145, 2011.

\bibitem[Collins(1996)]{collins1996space}
R.~T. Collins.
\newblock A space-sweep approach to true multi-image matching.
\newblock In \emph{Conference on Computer Vision and Pattern Recognition
  (CVPR)}, 1996.

\bibitem[Cox et~al.(1997)Cox, Little, and O'Shea]{Cox97}
D.~A. Cox, J.~Little, and D.~O'Shea.
\newblock \emph{Ideals, varieties, and algorithms - an introduction to
  computational algebraic geometry and commutative algebra}.
\newblock Springer, 1997.
\newblock ISBN 978-0-387-94680-1.

\bibitem[Fischler and Bolles(1981)]{Fischler81}
M.~A. Fischler and R.~C. Bolles.
\newblock Random sample consensus: a paradigm for model fitting with
  applications to image analysis and automated cartography.
\newblock In \emph{Communications of the ACM}, 1981.

\bibitem[Furgale et~al.(2013)Furgale, Schwesinger, Rufli, Derendarz, Grimmett,
  M\"{u}hlfellner, Wonneberger, Rottmann, Li, Schmidt, Nguyen, Cardarelli,
  Cattani, Br\"{u}ning, Horstmann, Stellmacher, Mielenz, K\"{o}ser, Beermann,
  H\"{a}ne, Heng, Lee, Fraundorfer, Iser, Triebel, Posner, Newman, Wolf,
  Pollefeys, Brosig, Effertz, Pradalier, and Siegwart]{FurgaleIV2013}
P.~Furgale, U.~Schwesinger, M.~Rufli, W.~Derendarz, H.~Grimmett,
  P.~M\"{u}hlfellner, S.~Wonneberger, J.~T.~S. Rottmann, B.~Li, B.~Schmidt,
  T.~N. Nguyen, E.~Cardarelli, S.~Cattani, S.~Br\"{u}ning, S.~Horstmann,
  M.~Stellmacher, H.~Mielenz, K.~K\"{o}ser, M.~Beermann, C.~H\"{a}ne, L.~Heng,
  G.~H. Lee, F.~Fraundorfer, R.~Iser, R.~Triebel, I.~Posner, P.~Newman,
  L.~Wolf, M.~Pollefeys, S.~Brosig, J.~Effertz, C.~Pradalier, and R.~Siegwart.
\newblock {Toward Automated Driving in Cities using Close-to-Market Sensors, an
  Overview of the V-Charge Project}.
\newblock In \emph{IEEE Intelligent Vehicles Symposium (IV)}, pages 809--816,
  Gold Coast, Australia, 23--26 June 2013.

\bibitem[Gallup et~al.(2007)Gallup, Frahm, Mordohai, Yang, and
  Pollefeys]{gallup2007real}
D.~Gallup, J.-M. Frahm, P.~Mordohai, Q.~Yang, and M.~Pollefeys.
\newblock Real-time plane-sweeping stereo with multiple sweeping directions.
\newblock In \emph{Conference on Computer Vision and Pattern Recognition
  (CVPR)}, 2007.

\bibitem[Gallup et~al.(2010)Gallup, Frahm, and Pollefeys]{gallup2010heightmap}
D.~Gallup, J.-M. Frahm, and M.~Pollefeys.
\newblock A heightmap model for efficient 3d reconstruction from street-level
  video.
\newblock In \emph{International Conference on 3D Data Processing,
  Visualization and Transmission (3DPVT)}, 2010.

\bibitem[Guo et~al.(2012)Guo, Mirzaei, and Roumeliotis]{guo12}
C.~Guo, F.~Mirzaei, and S.~Roumeliotis.
\newblock An analytical least-squares solution to the odometer-camera extrinsic
  calibration problem.
\newblock In \emph{IEEE International Conference on Robotics and Automation
  (ICRA)}, pages 3962--3968, 2012.

\bibitem[H{\"a}ne et~al.(2011)H{\"a}ne, Zach, Lim, Ranganathan, and
  Pollefeys]{hane2011stereo}
C.~H{\"a}ne, C.~Zach, J.~Lim, A.~Ranganathan, and M.~Pollefeys.
\newblock Stereo depth map fusion for robot navigation.
\newblock In \emph{IEEE/RSJ International Conference on Intelligent Robots and
  Systems (IROS)}, 2011.

\bibitem[H{\"a}ne et~al.(2014)H{\"a}ne, Heng, Lee, Sizov, and
  Pollefeys]{hane2014real}
C.~H{\"a}ne, L.~Heng, G.~H. Lee, A.~Sizov, and M.~Pollefeys.
\newblock Real-time direct dense matching on fisheye images using
  plane-sweeping stereo.
\newblock In \emph{International Conference on 3D Vision (3DV)}, 2014.

\bibitem[H{\"a}ne et~al.(2015)H{\"a}ne, Sattler, and
  Pollefeys]{hane2015obstacle}
C.~H{\"a}ne, T.~Sattler, and M.~Pollefeys.
\newblock Obstacle detection for self-driving cars using only monocular cameras
  and wheel odometry.
\newblock In \emph{IEEE/RSJ International Conference on Intelligent Robots and
  Systems (IROS)}, 2015.

\bibitem[Haralick et~al.(1991)Haralick, Lee, Ottenburg, and
  Nolle]{haralick1991pose}
R.~Haralick, D.~Lee, K.~Ottenburg, and M.~Nolle.
\newblock Analysis and solutions of the three point perspective pose estimation
  problem.
\newblock In \emph{IEEE Conference on Computer Vision and Pattern Recognition
  (CVPR)}, 1991.

\bibitem[Hartley and Zisserman(2004)]{Hartley2004}
R.~I. Hartley and A.~Zisserman.
\newblock \emph{Multiple View Geometry in Computer Vision}.
\newblock Cambridge University Press, ISBN: 0521540518, second edition, 2004.

\bibitem[Heng et~al.(2014)Heng, B{\"u}rki, Lee, Furgale, Siegwart, and
  Pollefeys]{Heng2014ICRA}
L.~Heng, M.~B{\"u}rki, G.~H. Lee, P.~Furgale, R.~Siegwart, and M.~Pollefeys.
\newblock Infrastructure-based calibration of a multi-camera rig.
\newblock In \emph{IEEE International Conference on Robotics and Automation
  (ICRA)}, 2014.

\bibitem[Heng et~al.(2015)Heng, Furgale, and Pollefeys]{Heng2015JFR}
L.~Heng, P.~Furgale, and M.~Pollefeys.
\newblock Leveraging image-based localization for infrastructure-based
  calibration of a multi-camera rig.
\newblock \emph{Journal of Field Robotics (JFR)}, 32:\penalty0 775--802, 2015.

\bibitem[Inc.({\natexlab{a}})]{google-selfdriving-car}
A.~Inc.
\newblock Google self-driving car project.
\newblock \url{https://www.google.com/selfdrivingcar}, {\natexlab{a}}.

\bibitem[Inc.({\natexlab{b}})]{tesla-autopilot}
T.~M. Inc.
\newblock Tesla autopilot.
\newblock \url{https://www.teslamotors.com/blog/your-autopilot-has-arrived},
  {\natexlab{b}}.

\bibitem[Kang et~al.(2001)Kang, Szeliski, and Chai]{kang2001handling}
S.~B. Kang, R.~Szeliski, and J.~Chai.
\newblock Handling occlusions in dense multi-view stereo.
\newblock In \emph{IEEE Computer Society Conference on Computer Vision and
  Pattern Recognition (CVPR)}, 2001.

\bibitem[Kumar et~al.(2008)Kumar, Ilie, Frahm, and Pollefeys]{kumar08}
R.~Kumar, A.~Ilie, J.~Frahm, and M.~Pollefeys.
\newblock Simple calibration of non-overlapping cameras with a mirror.
\newblock In \emph{IEEE Conference on Computer Vision and Pattern Recognition
  (CVPR)}, pages 1--7, 2008.

\bibitem[Lebraly et~al.(2011)Lebraly, Royer, Ait-Aider, Deymier, and
  Dhome]{lebraly11}
P.~Lebraly, E.~Royer, O.~Ait-Aider, C.~Deymier, and M.~Dhome.
\newblock Fast calibration of embedded non-overlapping cameras.
\newblock In \emph{IEEE International Conference on Robotics and Automation
  (ICRA)}, pages 221--227, 2011.

\bibitem[Lee et~al.(2013{\natexlab{a}})Lee, Fraundorfer, and
  Pollefeys]{IROS13_Lee_Robust}
G.~H. Lee, F.~Fraundorfer, and M.~Pollefeys.
\newblock Robust pose-graph loop-closures with expectation-maximization.
\newblock In \emph{IEEE/RSJ International Conference on Intelligent Robots and
  Systems (IROS)}, 2013{\natexlab{a}}.

\bibitem[Lee et~al.(2013{\natexlab{b}})Lee, Fraundorfer, and
  Pollefeys]{IROS13_Lee_Structureless}
G.~H. Lee, F.~Fraundorfer, and M.~Pollefeys.
\newblock Structureless pose-graph loop-closure with a multi-camera system on a
  self-driving car.
\newblock In \emph{IEEE/RSJ International Conference on Intelligent Robots and
  Systems (IROS)}, 2013{\natexlab{b}}.

\bibitem[Lee et~al.(2013{\natexlab{c}})Lee, Fraundorfer, and
  Pollefeys]{leeCVPR13}
G.~H. Lee, F.~Fraundorfer, and M.~Pollefeys.
\newblock Motion estimation for a self-driving car with a generalized camera.
\newblock In \emph{IEEE Conference on Computer Vision and Pattern Recognition
  (CVPR)}, 2013{\natexlab{c}}.

\bibitem[Lee et~al.(2013{\natexlab{d}})Lee, Li, Pollefeys, and
  Fraundorfer]{leeISRR13}
G.~H. Lee, B.~Li, M.~Pollefeys, and F.~Fraundorfer.
\newblock Minimal solutions for pose estimation of a multi-camera system.
\newblock In \emph{International Symposium on Robotics Research (ISRR)},
  2013{\natexlab{d}}.

\bibitem[Lee et~al.(2014{\natexlab{a}})Lee, , and Pollefeys]{LeeICRA14}
G.~H. Lee, , and M.~Pollefeys.
\newblock Unsupervised learning of threshold for geometric verification in
  visual-based loop-closure.
\newblock In \emph{International Conference on Robotics and Automation (ICRA)},
  2014{\natexlab{a}}.

\bibitem[Lee et~al.(2014{\natexlab{b}})Lee, Pollefeys, and
  Fraundorfer]{Lee14CVPR}
G.~H. Lee, M.~Pollefeys, and F.~Fraundorfer.
\newblock Relative pose estimation for a multi-camera system with known
  vertical direction.
\newblock In \emph{IEEE Conference on Computer Vision and Pattern Recognition
  (CVPR)}, 2014{\natexlab{b}}.

\bibitem[Lee et~al.(2015)Lee, Li, Pollefeys, and Fraundorfer]{leeIJRR15}
G.~H. Lee, B.~Li, M.~Pollefeys, and F.~Fraundorfer.
\newblock Minimal solutions for the multi-camera pose estimation problem.
\newblock In \emph{The International Journal of Robotics Research (IJRR)},
  2015.

\bibitem[Levinson and Thrun(2010)]{LevinsonICRA2010}
J.~Levinson and S.~Thrun.
\newblock Robust vehicle localization in urban environments using probabilistic
  maps.
\newblock In \emph{IEEE International Conference on Robotics and Automation
  (ICRA)}, 2010.

\bibitem[Li et~al.(2013)Li, Heng, K\"{o}ser, and Pollefeys]{li13}
B.~Li, L.~Heng, K.~K\"{o}ser, and M.~Pollefeys.
\newblock A multiple-camera system calibration toolbox using a feature
  descriptor-based calibration pattern.
\newblock In \emph{IEEE/RSJ International Conference on Intelligent Robots and
  Systems (IROS)}, pages 1301--1307, 2013.

\bibitem[Li et~al.(2008)Li, Hartley, and Kim]{Li08}
H.~D. Li, R.~Hartley, and J.~H. Kim.
\newblock A linear approach to motion estimation using generalized camera
  models.
\newblock In \emph{IEEE Conference on Vision and Pattern Recognition (CVPR)},
  2008.

\bibitem[Mei and Rives(2007)]{Mei2007ICRA}
C.~Mei and P.~Rives.
\newblock Single view point omnidirectional camera calibration from planar
  grids.
\newblock In \emph{IEEE International Conference on Robotics and Automation
  (ICRA)}, 2007.

\bibitem[Moreno-Noguer et~al.(2007)Moreno-Noguer, Lepetit, and Fua]{Moreno07}
F.~Moreno-Noguer, V.~Lepetit, and P.~Fua.
\newblock Accurate non-iterative o(n) solution to the pnp problem.
\newblock In \emph{International Conference on Computer Vision (ICCV)}, 2007.

\bibitem[Nist\'er and Stew\'enius(2006)]{Nister06}
D.~Nist\'er and H.~Stew\'enius.
\newblock Scalable recognition with a vocabulary tree.
\newblock In \emph{IEEE Conference on Computer Vision and Pattern Recognition
  (CVPR)}, 2006.

\bibitem[Pless(2003)]{Pless03}
R.~Pless.
\newblock Using many cameras as one.
\newblock In \emph{IEEE Conference on Vision and Pattern Recognition (CVPR)},
  2003.

\bibitem[Quan and Lan(1999)]{Long99}
L.~Quan and Z.~D. Lan.
\newblock Linear n-point camera pose determination.
\newblock In \emph{Pattern Analysis and Machine Intelligence}, 1999.

\bibitem[Sattler et~al.(2011)Sattler, Leibe, and Kobbelt]{sattler11}
T.~Sattler, B.~Leibe, and L.~Kobbelt.
\newblock Fast image-based localization using direct 2d-to-3d matching.
\newblock In \emph{IEEE International Conference on Computer Vision (ICCV)},
  2011.

\bibitem[Siegwart et~al.(2011)Siegwart, Nourbakhsh, and Scaramuzza]{Siegwart11}
R.~Siegwart, I.~Nourbakhsh, and D.~Scaramuzza.
\newblock \emph{Introduction to Autonomous Mobile Robots}.
\newblock MIT Press, 2nd edition, 2011.

\bibitem[Stew{\'e}nius et~al.(2005)Stew{\'e}nius, Nist{\'e}r, Oskarsson, and
  {\AA}str{\"o}m]{Stewenius05}
H.~Stew{\'e}nius, D.~Nist{\'e}r, M.~Oskarsson, and K.~{\AA}str{\"o}m.
\newblock Solutions to minimal generalized relative pose problems.
\newblock In \emph{Workshop on Omnidirectional Vision and Camera Networks
  (OMNIVIS)}, 2005.

\bibitem[Tian and Huhns(1986)]{tian1986algorithms}
Q.~Tian and M.~N. Huhns.
\newblock Algorithms for subpixel registration.
\newblock \emph{Computer Vision, Graphics, and Image Processing}, 35\penalty0
  (2):\penalty0 220--233, 1986.

\bibitem[Yang and Pollefeys(2003)]{yang2003multi}
R.~Yang and M.~Pollefeys.
\newblock Multi-resolution real-time stereo on commodity graphics hardware.
\newblock In \emph{Conference on Computer Vision and Pattern Recognition
  (CVPR)}, 2003.

\bibitem[Ziegler et~al.(2014)Ziegler, Bender, Schreiber, Lategahn, Strauss,
  Stiller, Dang, Franke, Appenrodt, Keller, Kaus, Herrtwich, Rabe, Pfeiffer,
  Lindner, Stein, Erbs, Enzweiler, Knoppel, Hipp, Haueis, Trepte, Brenk, Tamke,
  Ghanaat, Braun, Joos, Fritz, Mock, Hein, and Zeeb]{ZieglerITSM2014}
J.~Ziegler, P.~Bender, M.~Schreiber, H.~Lategahn, T.~Strauss, C.~Stiller,
  T.~Dang, U.~Franke, N.~Appenrodt, C.~G. Keller, E.~Kaus, R.~G. Herrtwich,
  C.~Rabe, D.~Pfeiffer, F.~Lindner, F.~Stein, F.~Erbs, M.~Enzweiler,
  C.~Knoppel, J.~Hipp, M.~Haueis, M.~Trepte, C.~Brenk, A.~Tamke, M.~Ghanaat,
  M.~Braun, A.~Joos, H.~Fritz, H.~Mock, M.~Hein, and E.~Zeeb.
\newblock Making bertha drive - an autonomous journey on a historic route.
\newblock \emph{IEEE Intelligent Transportation Systems Magazine}, 6\penalty0
  (2):\penalty0 8--20, 2014.

\end{thebibliography}

\end{document}